\documentclass{article}

\usepackage{iclr2027_conference,times}

\usepackage{amsmath,amsfonts,bm}
\def\eqref#1{equation~\ref{#1}}
\def\1{\bm{1}}

\DeclareMathAlphabet{\mathsfit}{\encodingdefault}{\sfdefault}{m}{sl}
\SetMathAlphabet{\mathsfit}{bold}{\encodingdefault}{\sfdefault}{bx}{n}

\usepackage{amssymb}

\usepackage{graphicx}
\usepackage{wrapfig}
\usepackage{needspace}
\usepackage{booktabs}
\usepackage{tabularx}
\usepackage{colortbl}
\usepackage{multirow}
\usepackage{arydshln}
\usepackage{enumitem}
\usepackage{algorithm}
\usepackage{algpseudocode}
\usepackage{hyperref}
\usepackage{url}

\usepackage[most]{tcolorbox}

\newcommand{\wsrdrop}[1]{{\scriptsize\color{red}($\downarrow #1$)}}

\newtcolorbox{finding}[1]{
  colback=gray!10,
  colframe=black,
  boxrule=0.8pt,
  arc=2mm,
  left=3mm,
  right=3mm,
  top=1.7mm,
  bottom=1.7mm,
  before upper={\textbf{\textit{Finding}~#1.}\enspace}
}

\definecolor{imageinputbg}{HTML}{E6F3FF}
\definecolor{stateinputbg}{HTML}{E7F8F2}
\definecolor{bothinputbg}{HTML}{F4EAFF}

\title{Are Vision-Language-Action Models Robust to One-Step Observation Perturbations?}
\author{
Shojiro Yamabe\\
Science Tokyo\\
\texttt{yamabe.s.2fb0@m.isct.ac.jp}
\And
Jun Sakuma\\
Science Tokyo, RIKEN
}

\iclrfinalcopy

\hypersetup{
  hidelinks,
  pdftitle={Are Vision-Language-Action Models Robust to One-Step Observation Perturbations?},
  pdfauthor={Shojiro Yamabe and Jun Sakuma}
}

\begin{document}

\maketitle
% The conference style sets its publication header inside \maketitle.
\lhead{Preprint}
% !TeX root = preprint.tex

% Abstract and main-paper sections.
% \begin{abstract}
% Ensuring the safety of vision-language-action (VLA) models is essential for their deployment in the physical world. One safety risk is momentary observation corruption caused by communication failures or blur.
% However, prior work has mainly considered persistent perturbations.
% In this work, we evaluate robustness to one-step perturbations applied at a single inference step. 
% Our experiments show that these perturbations substantially degrade VLA performance and that their impact depends on the action chunk execution length. 
% Based on these findings, we propose CARE, which dynamically selects the execution length based on agreement with the previously predicted action chunk. 
% CARE improves robustness with low computational overhead while preserving clean performance by selecting shorter execution lengths only under perturbations. More broadly, our work introduces a temporal perspective on perturbations for VLA safety research.
% \end{abstract}

\begin{abstract}
Understanding the safety risks of vision-language-action (VLA) models is essential for their deployment in the physical world.
Existing safety research has mainly considered persistent perturbations that are applied continuously to observations throughout an episode.
However, momentary observation corruption, in which observations are severely perturbed only briefly within an episode, remains an underexplored safety threat.
To address this gap, this work investigates robustness to one-step perturbations applied at a single time step per episode.
Our experiments reveal that these perturbations substantially degrade VLA performance and that their impact depends on the action chunk execution length. 
Based on them, we propose CARE, which dynamically selects the execution length based on consistency with the previously predicted action chunk. 
CARE improves robustness with low computational overhead while preserving clean performance by selecting shorter execution lengths only under perturbations.
% Our work introduces a temporal perspective for VLA safety research.
\end{abstract}

\section{Introduction}\label{sec:introduction}
% Vision–Language–Action（VLA）モデルは，視覚と言語の理解能力を，ロボットによる物理的な行動の生成へと拡張する基盤モデルである~\citep{kim2024openvla,black2024pi_0}．
% ロボットアーム~\citep{black2024pi_0,black2025pi}をはじめとする，実世界で動作するロボットへの応用が進められている．一方，こうした環境ではVLAの出力が物理世界に直接作用するため，誤った行動はタスクの失敗にとどまらず，周囲の物体や人への損害につながる可能性がある．
% VLAを実環境に安全に導入するためには，物理世界におけるさまざまなノイズに対する頑健性を明らかにすることが重要である．
% Vision-Language-Action (VLA) models are foundation models that extend visual and linguistic understanding to robotic action generation~\citep{kim2024openvla,black2024pi_0}.
% VLAs are increasingly being applied to real-world robots, including robotic arms~\citep{black2024pi_0,black2025pi}.
% Because these models directly control physical systems, incorrect actions can cause task failure, damage nearby objects, or harm people.
% Safe deployment of VLAs therefore requires understanding their robustness to real-world perturbations.
Vision-Language-Action (VLA) models provide a foundation for robot control by extending visual and linguistic understanding to action generation~\citep{kim2024openvla,black2024pi_0}.
% VLAs are increasingly applied to real-world robotic systems, including robotic arms~\citep{black2024pi_0,black2025pi}.
Since VLA outputs are executed as physical actions, inappropriate actions can not only cause task failure, but also damage nearby objects or harm people.
Safe deployment therefore requires VLAs to remain reliable under observation degradation that arises in real-world environments.

One important safety concern is momentary observation corruption, where observations are severely degraded for only a brief period.
For example, sudden illumination changes can temporarily overexpose a camera, and brief communication failures can lead to abnormal sensor values.
% Although these corruptions may seem to pose little risk because they are brief, they can cause irreversible damage if they occur at critical moments, such as dropping and breaking an object, or leave objects out of the robot's reach.
While seemingly innocuous, these corruptions can cause irreversible damage when they occur at critical moments, such as dropping and breaking an object or leaving it out of the robot's reach.
As a result, the robot may be unable to resume the task even after reliable observations are restored.
However, robustness to such corruptions has received limited attention in prior work~\citep{guo2026on,wang2025vlatest,hancock2025run,zhang2025gevrm}.
Existing evaluations typically consider \emph{persistent perturbations} which are applied continuously throughout an episode, leaving robustness to momentary observation corruption poorly understood.

\begin{figure}[t]
    \centering
    \includegraphics[width=\linewidth]{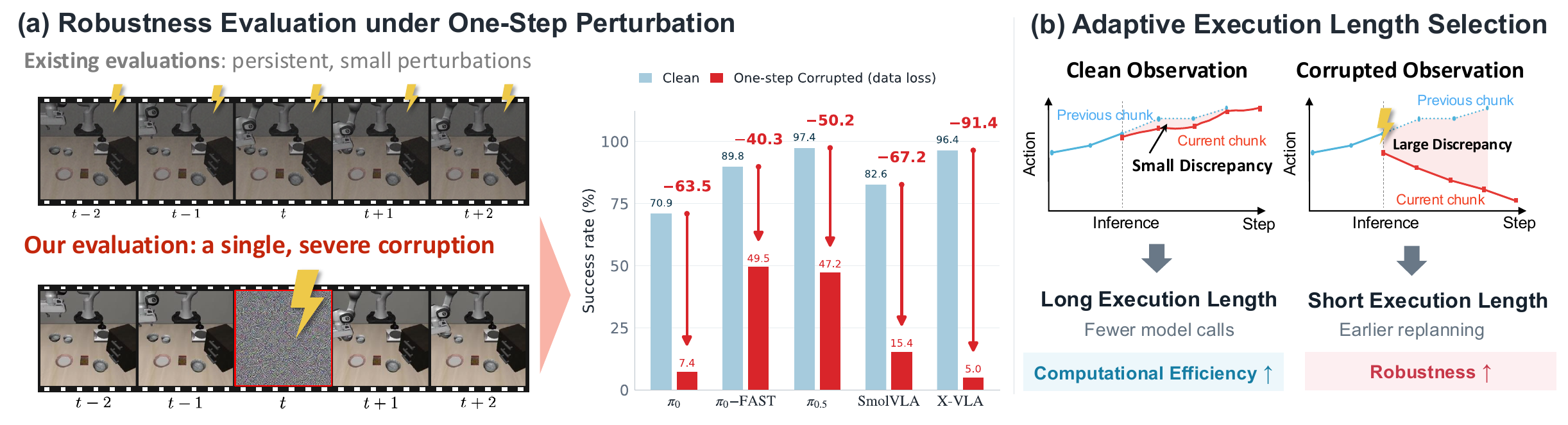}
    \caption{\textbf{Overview.} (a) We evaluate VLA robustness to one-step perturbations and observe substantial success rate drops across 40 LIBERO tasks. Corrupted scores average the lowest 10\% of success rates across evaluated intervention steps. (b) Adaptive execution length selection based on prediction consistency shortens execution under perturbations to improve robustness.}
    \label{fig:one_step_overview}
    \vspace{-12pt}
\end{figure}

In this work, we study the robustness of VLA models to \emph{one-step perturbations}, each applied at exactly one step in an episode (Figure~\ref{fig:one_step_overview} (a)).
We design eight types of perturbations that emulate momentary observation corruptions and systematically evaluate their effects across five VLA models.
For each perturbation, we evaluate every inference step as a candidate intervention time in separate rollouts, with each episode receiving exactly one perturbed observation.
Our experiments reveal two important findings about the robustness of current VLAs:

% \vspace{-0.5em}
% \paragraph{Finding 1: A single corrupted observation causes substantial performance degradation across evaluated VLAs.}
% The impact of a one-step perturbation varies substantially with its timing.
% To assess performance at particularly vulnerable steps, we average the lowest 10\% of success rates across evaluated intervention steps.
% Under this metric, every evaluated model exhibits a drop of more than 50 percentage points relative to its clean success rate under at least one perturbation.
% This suggests a vulnerability in which the effect of a one-step perturbation, which corresponds to approximately 0.03 seconds for a policy running at 30 Hz, can persist beyond the perturbed step.

\vspace{-0.5em}
\paragraph{Finding 1: One-step perturbations at critical moments cause substantial performance degradation across evaluated VLAs.}
The impact of a one-step perturbation varies substantially depending on when the perturbation occurs.
To characterize performance at particularly vulnerable intervention steps, we report the average of the lowest 10\% of success rates across evaluated steps.
Under this metric, every evaluated model exhibits a drop of more than 50 percentage points relative to its clean success rate under at least one perturbation.
% These results suggest that even a one-step perturbation, which corresponds to approximately 0.03 seconds for a policy running at 30 Hz, can cause substantial performance degradation when it occurs at a critical moment.
These results suggest that even a single corrupted observation can cause substantial performance degradation when it occurs at a critical moment.

% 第二に，この脆弱性を生む要因の一つが，VLAで広く用いられているアクションチャンキング（action chunking）~\citep{zhao2023learning}であることを明らかにした．
% アクションチャンキングとは，1回の推論で複数ステップ分の行動列をまとめて生成し，そのうち，事前に定めた実行長分の行動を順次実行する推論機構である．
% この機構は，短い制御周期でのロボット制御を可能にする一方，誤った行動列が生成されると，その行動が実行長にわたって継続するため，後続の振る舞いを不安定にする可能性がある．
% 実行長を短く設定すればこの脆弱性を改善できるものの，計算コストの増加や行動間の整合性の低下により，タスク性能と頑健性の両立が困難になる~\citep{chi2025diffusion,black2026real}．
% Second, we show that action chunking~\citep{zhao2023learning}, a mechanism widely used in VLAs, contributes to this vulnerability.
% Action chunking predicts multiple future actions in a single inference call and executes a fixed number of them before the next call. We refer to this number as the execution length.
% This mechanism supports high-frequency robot control. However, if the model predicts an incorrect action chunk, the robot continues to execute its actions throughout the execution length, potentially destabilizing subsequent behavior.
% Shortening the execution length can mitigate this vulnerability, but it increases computational cost and can reduce action consistency, making it difficult to maintain both task performance and robustness~\citep{chi2025diffusion,black2026real}.

\vspace{-0.5em}
\paragraph{Finding 2: Vulnerability to one-step perturbations depends on the action execution length.}
Action chunking~\citep{zhao2023learning} is widely used in VLAs to predict several future actions at once.
The robot then executes a fixed number of predicted actions before the next prediction, which we call the execution length.
If a perturbation causes the model to predict an incorrect action chunk, the robot continues executing actions from that chunk until the next prediction.
We find that shortening the execution length reduces this vulnerability.
However, a shorter execution length requires more frequent inference and may reduce action consistency, creating a trade-off between robustness and task performance~\citep{chi2025diffusion,black2026real}.

% この知見に基づき，本研究では，予測された行動列の整合性に応じて実行長を動的に調整する，新たなアクションチャンキング手法を提案する．
% 本手法では，前回の推論で生成された行動列のうち未実行の行動と，今回の推論で生成された行動列との差分を計算する．
% 両者の差分が大きい場合は，現在の観測または行動予測が不安定である可能性が高いと判断する．そこで，短い実行長を選択することにより，誤った行動列が長時間にわたって実行されることを防ぐ．
% 一方，差分が小さい場合は，長い実行長を選択することにより，モデルの推論回数を削減する．
% 提案手法は，追加のモデル推論や学習を必要とせず，行動列を出力する幅広いVLAモデルに適用できる．
% 実験により，提案手法が通常時のタスク性能と推論効率を維持しながら，単一時刻の入力摂動に対する頑健性を大幅に向上させることを示す．

% Motivated by these findings, we propose \textit{CARE (Consistency-Aware Robust Execution Length Selection)}, an adaptive execution length selection method based on the consistency of consecutive action predictions.
% CARE uses action chunks that extend beyond the actions to be executed, retaining the unexecuted portion for a consistency check at the next inference step.
% At each subsequent inference step, it compares the newly predicted actions with the retained predictions over the same future timesteps.
% A large discrepancy suggests that the current observation or action prediction may be unstable.
% In this case, CARE selects a short execution length to limit the prolonged execution of potentially erroneous actions.
% When the discrepancy is small, it selects a long execution length to reduce the number of model calls.
% For VLA models that already predict beyond their execution length, CARE reuses the existing outputs without modifying the model or requiring additional model calls or training.
% Experiments show that CARE substantially improves robustness to one-step input perturbations while maintaining clean task performance and inference efficiency.

Motivated by these findings, we propose Consistency-Aware Robust Execution length selection (CARE), which adjusts the execution length based on the consistency of consecutive action predictions (Figure~\ref{fig:one_step_overview} (b)).
At each inference step, our method compares the newly predicted action sequence with the unexecuted portion of the previous prediction.
A large difference suggests that the current observation or action prediction may be unstable. In this case, the method selects a short execution length to prevent an incorrect action sequence from being executed for an extended period.
This simple design requires no additional model calls or training and is broadly applicable to VLA models that predict action chunks.
Experiments show that CARE substantially improves robustness to one-step input perturbations with low computational overhead, while largely preserving clean task performance for most evaluated models.

% 本研究の貢献は，以下の4点である．
% Our contributions are summarized as follows:
% \begin{itemize}
% % \item 一時的な観測破損により，現在のVLAのタスク性能が大幅に低下し得るという脆弱性を明らかにした．
% % \item この脆弱性を分析し，アクションチャンキングが入力摂動の影響を後続時刻まで持続させることが，脆弱性を増幅する要因の一つであることを示した．
% % \item この脆弱性への対策として，実行長を動的に調整する新たなアクションチャンキング手法を提案した．我々の知る限り，安全性の観点からアクションチャンキングの実行長を動的に調整する手法を提案するのは，本研究が初めてである．
% % \item 実験により，提案手法が通常時のタスク性能と推論効率を維持しながら，頑健性を向上させることを示した．
% \item We reveal a vulnerability in current VLAs whereby transient observation corruption can substantially degrade task performance.
% \item We analyze this vulnerability and show that action chunking amplifies it by allowing the effects of input perturbations to persist into subsequent steps.
% \item To address this vulnerability, we propose a new adaptive execution length selection method. To the best of our knowledge, this is the first work to dynamically adapt the execution length of action chunking to improve safety.
% \item We experimentally show that the proposed method improves robustness while maintaining clean task performance and inference efficiency.`'
% \end{itemize}

Our contributions are summarized as follows:
\begin{itemize}[leftmargin=*, topsep=0pt, itemsep=2pt, parsep=0pt]
\item We present the first systematic study of VLA robustness to one-step perturbations and reveal severe task-performance degradation across several VLA models.
\item We analyze this vulnerability and identify longer action-chunk execution lengths as a key factor that amplifies the effect of a one-step perturbation.
\item We propose CARE, a training-free and broadly applicable method for adaptive execution length selection, and demonstrate its effectiveness across multiple models and benchmarks.
\end{itemize}

\section{Related Work}\label{sec:related_work}
This section provides an overview of related work, with further details in Appendix~\ref{app:related_work}.

% \vspace{-8pt}
% \paragraph{Vision-Language-Action Models.}
% VLA models extend the visual and linguistic understanding capabilities of Vision-Language Models (VLMs), trained on large-scale vision-language datasets, to robotic action generation~\citep{brohan2023rt,zitkovich2023rt,mees2024octo,kim2024openvla,black2024pi_0,black2025pi,pertsch2025fast,shukor2025smolvla,wen2025dexvla,zheng2026xvla,liu2025rdt}.
% These models are increasingly being applied to robots operating in the physical world, including robotic arms and mobile robots~\citep{black2024pi_0,team2025gemini}.
% This work focuses on the safety of VLA models, examining their robustness to observation corruption during real-world operation.

\vspace{-8pt}
\paragraph{Robustness to Input Perturbations.}
A growing body of work has investigated VLA safety to support safe deployment in real-world environments~\citep{zhang2025badrobot,robey2025jailbreaking,jones2025adversarial,li2026vision}.
One focus of this research is robustness to input perturbations during inference.
Studies in this area have evaluated VLA robustness to perturbations affecting image inputs, state inputs, or both, reporting substantial drops in task success rates~\citep{guo2026on,wang2025vlatest,hancock2025run,wang2025exploring,lu2026phantom,xie2026strong,zhang2025robustvla,xu2025model,li2025cronusvla}.
However, these studies mainly assume persistent perturbations, which are continuously applied throughout an episode.
This work evaluates the robustness against one-step perturbations that intervene at only a single step.

\vspace{-8pt}
\paragraph{Action Chunking.}
Action chunking~\citep{zhao2023learning,chi2025diffusion} predicts multiple future actions in a single inference pass and sequentially executes a fixed number of them.
While this mechanism enables high-frequency robot control, it reduces responsiveness to rapid changes in motion.
To address this trade-off, several methods have been proposed to dynamically adjust the execution length~\citep{wang2026trust,liang2026adaptive,pan2026vla,wang2026vla,jing2025mixture}.
However, these methods do not consider perturbations to the observations, limiting their contribution to robustness.
In this work, we propose CARE, an adaptive execution length selection method that uses the consistency with the predicted action chunk to improve robustness.
While prior work has used temporal consistency across action predictions for other purposes~\citep{pmlr-v270-agia25a,liu2025bidirectional,so2025improving}, CARE first uses it for adaptive execution length selection.

\section{Preliminaries}\label{sec:preliminaries}
We consider a robotic control setting in which a VLA policy makes sequential decisions to control a robot to accomplish a goal specified by a language instruction.
At step $t$, the policy $\pi_\theta$ takes three inputs as observation: an image input $v_t$ consisting of RGB images; a robot state $s_t$, such as the end-effector pose and gripper state; and a language instruction $l$.
The policy then outputs an action chunk consisting of $K$ future actions:
\begin{equation}
    \mathbf{A}_{t}
    =
    (a_t, a_{t+1}, \ldots, a_{t+K-1})
    \sim
    \pi_\theta
    \left(
    \cdot
    \mid
    v_t, s_t, l
    \right),
\end{equation}
where $a_t$ denotes the action for step $t$. 
The robot executes only the first $H \leq K$ actions $(a_t, \ldots, a_{t+H-1})$ of the predicted chunk before the next policy inference. We refer to $H$ as the execution length.
After these $H$ actions have been executed, the policy generates the next action chunk $\mathbf{A}_{t+H}$ from $v_{t+H}$ and $s_{t+H}$.
% Thus, a momentary observation corruption can affect the policy output only when it coincides with an inference step.

% We restrict our attention to the common setting among recent VLAs in which policies condition only on the current observation, excluding models that incorporate a short observation history~\citep{brohan2023rt,mees2024octo}. Accordingly, if the observation is corrupted at only one inference step, subsequent policy inferences no longer directly receive the corrupted observation.

\section{Robustness Evaluation to One-Step Perturbations}\label{sec:robustness_evaluation}

We first ask whether momentary observation corruption can cause task failure in current VLAs.
Unlike persistent perturbations used in prior evaluations, real-world observation corruption caused by overexposure, motion blur, or sensor input loss may last only briefly.
To study this threat in a controlled setting, we consider one-step perturbations, which corrupt the observation at a single step while leaving observations at all other steps unchanged.
We evaluate robustness against these perturbations and investigate why its effects may persist beyond the perturbed step.

% また、前述のとおり、VLAモデルは実行長ごとに推論を行うため、推論時以外の観測はモデルの出力に影響しない。

% 本セクションでは、transient observation corruptionに対する頑健性を評価する。
% transient observation corruptionとはxxxである。
% 定量評価するために、現実の脅威を想定した摂動を設計して1stepにのみ加える。
% それにより、xxxとyyyであることを示す。

\subsection{Experimental Setup}\label{subsec:experimental_setup}
\vspace{-8pt}
\paragraph{Models and Benchmarks.}
% 評価対象のVLAモデルとして、$\pi_0$~\citep{black2024pi_0}、$\pi_0$-FAST~\citep{pertsch2025fast}、$\pi_{0.5}$~\citep{black2025pi}、SmolVLA~\citep{shukor2025smolvla}、およびX-VLA~\citep{zheng2026xvla}を用いる。
% 各モデルの実行長には、$H \in \{1, 5, 10, 15, 20, 25, 30\}$のうち、摂動を付与しない条件でタスク成功率が最も高かった値を採用する。具体的には、$\pi_0$、$\pi_{0.5}$、X-VLAでは$H=15$、$\pi_0$-FASTとSmolVLAでは$H=10$とする。
% シミュレーション環境には、ロボット操作ベンチマークLIBERO~\citep{liu2023libero}の4つのタスクスイート、LIBERO-Spatial、LIBERO-Object、LIBERO-Goal、およびLIBERO-Longを用いる。
% 実行長の選択に用いた実験結果などの詳細は、Appendix~\ref{app:experimental_details}に示す。
We evaluate five VLA models: $\pi_0$~\citep{black2024pi_0}, $\pi_0$-FAST~\citep{pertsch2025fast}, $\pi_{0.5}$~\citep{black2025pi}, SmolVLA~\citep{shukor2025smolvla}, and X-VLA~\citep{zheng2026xvla}.
For each model, we select the execution length from $H \in \{1, 5, 10, 15, 20, 25, 30\}$ that achieves the highest task success rate without perturbations. Specifically, we use $H=15$ for $\pi_0$, $\pi_{0.5}$, and X-VLA, and $H=10$ for $\pi_0$-FAST and SmolVLA.
As the simulation environments, we use four task suites from the LIBERO benchmark~\citep{liu2023libero}.
Further details are provided in Appendix~\ref{app:implementation_details}.

\vspace{-8pt}
\paragraph{Perturbations.}
\begin{figure}[t]
    \centering
    \includegraphics[width=\linewidth]{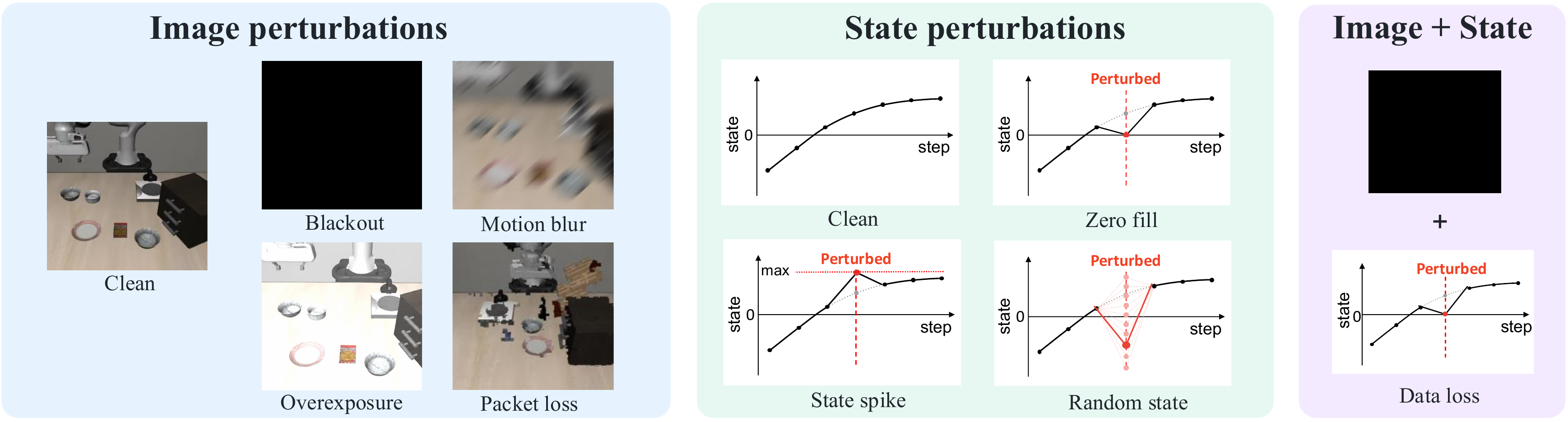}
    \vspace{-7mm}
    \caption{\textbf{Eight types of one-step perturbations.} Each perturbation corrupts the image input, the state input, or both at a single inference step.}
    \label{fig:perturbation_types}
    \vspace{-12pt}
\end{figure}
We consider eight perturbations that emulate momentary observation corruptions that may occur in real-world environments (Figure~\ref{fig:perturbation_types}).
We group them into three categories by input modality.
\textbf{(i) Image perturbations:}
These simulate disruptions to camera capture or image transmission.
We apply Blackout, Overexposure, Motion blur, and Packet loss.
\textbf{(ii) State perturbations:}
These simulate sensor malfunctions or communication failures.
We apply Zero fill, State spike, and Random state, which replace the affected state components with zero, their upper bounds, or random outliers, respectively.
\textbf{(iii) Image-and-state perturbations:}
This simulates a complete interruption of observation inputs.
We apply Data loss, which replaces the input image with a black image and all state components with zero.
Implementation details are provided in Appendix~\ref{app:perturbation_implementation}.

\vspace{-8pt}
\paragraph{Evaluation Protocol.}
We evaluate how task performance depends on when a perturbation is applied.
Since the policy receives observations only at inference steps, we evaluate each intervention step in separate rollouts, applying a perturbation at exactly one inference step per episode.
For example, when $H=10$, the perturbation is applied at a single inference step selected from $t=1,11,21,\ldots$.
For each model, task, and perturbation, we run 10 episodes per intervention step.
We also report \textbf{\emph{Bottom10\%-SR}} to assess performance at particularly vulnerable steps.
This metric is the mean of the lowest 10\% of success rates across evaluated intervention steps.

\subsection{Existing VLA Models are Vulnerable to One-Step Perturbations}\label{subsec:eval_vulnerability_finding1}

\begin{figure*}[t]
    \centering
    \includegraphics[width=\textwidth]{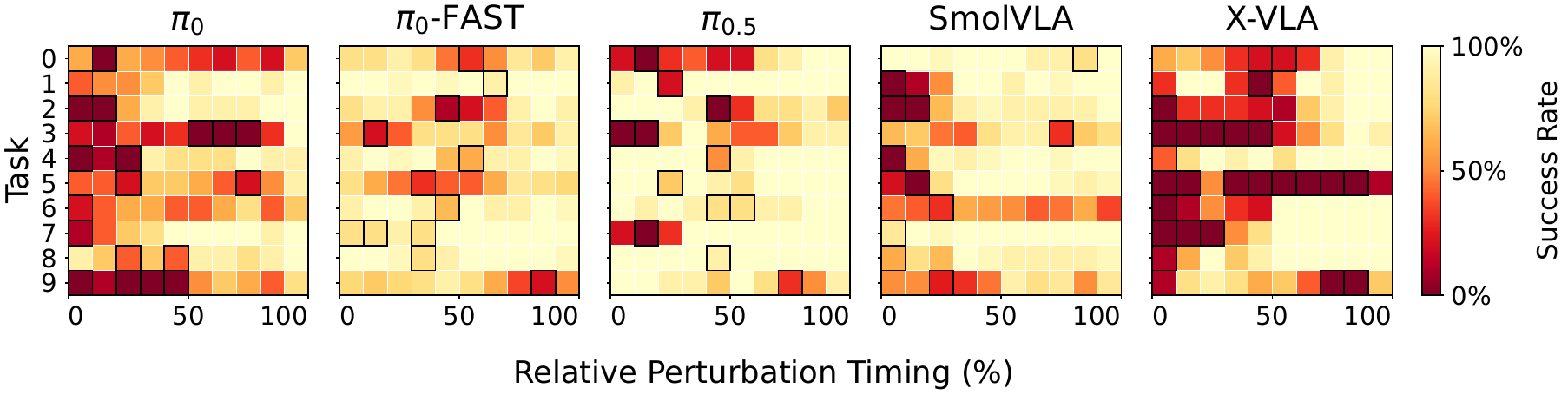}
    \vspace{-7mm}
    \caption{\textbf{Perturbation timing strongly affects task success.}
Heatmaps show success rates under data loss on LIBERO-Goal, with black boxes marking task-wise minima.}
    \label{fig:perturbation_timing_analysis_all_models}
    \vspace{-16pt}
\end{figure*}

\begin{table*}[t]
    \centering
    \caption{\textbf{Existing VLAs are vulnerable to one-step perturbations.}
    Clean success rate and Bottom10\%-SR under each perturbation (\%), averaged over 40 tasks on LIBERO.}
    \label{tab:natural_perturbation_robustness}
    \definecolor{imageinputbg}{HTML}{E6F3FF}
    \definecolor{stateinputbg}{HTML}{E7F8F2}
    \definecolor{bothinputbg}{HTML}{F4EAFF}
    \setlength{\tabcolsep}{3pt}
    \renewcommand{\arraystretch}{1.35}
    \resizebox{\textwidth}{!}{%
    \begin{tabular}{lc
        >{\columncolor{imageinputbg}}c
        >{\columncolor{imageinputbg}}c
        >{\columncolor{imageinputbg}}c
        >{\columncolor{imageinputbg}}c
        >{\columncolor{stateinputbg}}c
        >{\columncolor{stateinputbg}}c
        >{\columncolor{stateinputbg}}c
        >{\columncolor{bothinputbg}}c}
        \toprule
        \multirow{2}{*}{Model} & \multirow{2}{*}{Clean}
        & \multicolumn{4}{c}{\cellcolor{imageinputbg}Image input}
        & \multicolumn{3}{c}{\cellcolor{stateinputbg}State input}
        & \multicolumn{1}{c}{\cellcolor{bothinputbg}Image + state} \\
        \cmidrule(lr){3-6}\cmidrule(lr){7-9}\cmidrule(lr){10-10}
        & & Overexp. & Blackout & Blur & Packet loss & Zero fill & Spike & Random & Data loss \\
        \midrule
        $\pi_0$      & 70.9 & 53.9 \wsrdrop{17.0} & 48.1 \wsrdrop{22.8} & 51.4 \wsrdrop{19.5} & 50.2 \wsrdrop{20.7} & 5.9 \wsrdrop{65.0} & 13.8 \wsrdrop{57.1} & 19.6 \wsrdrop{51.3} & 7.4 \wsrdrop{63.5} \\
        $\pi_0$-FAST & 89.8 & 44.4 \wsrdrop{45.4} & 45.0 \wsrdrop{44.8} & 36.7 \wsrdrop{53.1} & 75.6 \wsrdrop{14.2} & 79.7 \wsrdrop{10.1} & 72.9 \wsrdrop{16.9} & 76.1 \wsrdrop{13.7} & 49.5 \wsrdrop{40.3} \\
        $\pi_{0.5}$  & 97.4 & 88.0 \wsrdrop{9.4} & 45.9 \wsrdrop{51.5} & 64.9 \wsrdrop{32.5} & 91.1 \wsrdrop{6.3} & 90.1 \wsrdrop{7.3} & 91.6 \wsrdrop{5.8} & 91.0 \wsrdrop{6.4} & 47.2 \wsrdrop{50.2} \\
        SmolVLA      & 82.6 & 33.3 \wsrdrop{49.3} & 34.8 \wsrdrop{47.8} & 44.1 \wsrdrop{38.5} & 69.5 \wsrdrop{13.1} & 53.3 \wsrdrop{29.3} & 52.3 \wsrdrop{30.3} & 58.5 \wsrdrop{24.1} & 15.4 \wsrdrop{67.2} \\
        X-VLA        & 96.4 & 90.5 \wsrdrop{5.9} & 28.6 \wsrdrop{67.8} & 38.5 \wsrdrop{57.9} & 41.5 \wsrdrop{54.9} & 15.0 \wsrdrop{81.4} & 2.8 \wsrdrop{93.6} & 10.4 \wsrdrop{86.0} & 5.0 \wsrdrop{91.4} \\
        \bottomrule
    \end{tabular}}
    \vspace{-12pt}
\end{table*}

\vspace{-8pt}
\paragraph{Perturbation timing strongly affects task success.}
Figure~\ref{fig:perturbation_timing_analysis_all_models} shows success rates under data loss across perturbation steps for five VLA models on LIBERO-Goal.
Relative perturbation timing is defined as the normalized position of the perturbed inference step within an episode.
Even for the same model and task, success rates differ substantially depending on when the perturbation is applied.
Although the most vulnerable steps differ across models and tasks, earlier perturbations tend to cause larger reductions in task success.
This pattern may arise from compounding errors, a well-known problem in imitation learning~\citep{ross2011reduction,xu2024humanvla}.
These findings suggest that robustness evaluations should account for particularly vulnerable steps. Additional results and discussion are provided in Appendix~\ref{app:perturbation_timing_analysis}.

\vspace{-8pt}
\paragraph{Robustness at vulnerable steps.}
We next compare model robustness at vulnerable perturbation steps using Bottom10\%-SR.
Table~\ref{tab:natural_perturbation_robustness} shows that every evaluated model has a Bottom10\%-SR more than 50 percentage points below its clean success rate under at least one perturbation.
These results reveal substantial vulnerability to one-step perturbations.
Furthermore, the perturbations that expose these vulnerabilities differ across models.
$\pi_0$-FAST and SmolVLA are particularly sensitive to image perturbations, whereas $\pi_0$ and X-VLA are especially vulnerable to state perturbations.

\subsection{Execution Length Amplifies Perturbation Effects}\label{subsec:analysis_action_chunk}
% Previous right-aligned Figure 1:
% \begin{wrapfigure}{r}{0.40\textwidth}
%     \centering
%     \includegraphics[width=\linewidth]{figures/execution_analysis.pdf}
%     \caption{Relationship between the execution length $H$ and the decrease in success rate. Smaller values indicate a smaller decrease in success rate due to a single-timestep perturbation.}
%     \label{fig:action_chunk_analysis}
% \end{wrapfigure}
\begin{figure*}[t]
    \centering
    \includegraphics[width=\textwidth]{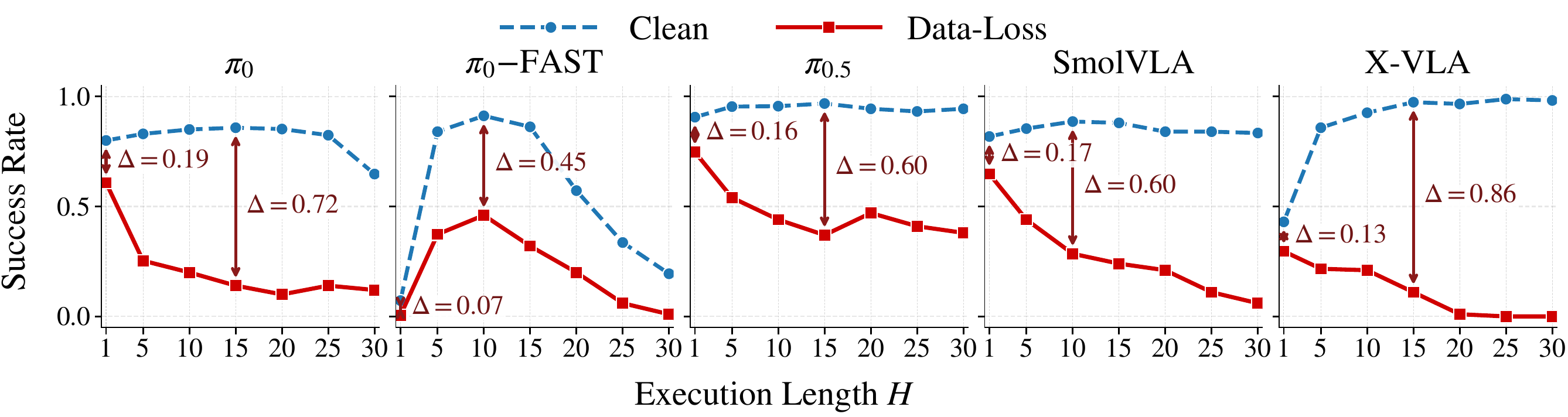}
    \vspace{-2em}
    \caption{\textbf{Execution Length Amplifies Perturbation Effects}. Clean success rate and Bottom10\%-SR under Data-loss perturbation across execution lengths $H$.}
    \label{fig:action_chunk_analysis}
    \vspace{-1em}
\end{figure*}

We hypothesize that action chunking contributes to the observed vulnerability to one-step perturbations.
A single perturbed observation can cause incorrect actions over multiple consecutive steps, making recovery more difficult.
To test this hypothesis, we compare robustness across execution lengths $H$. 
Because clean performance also depends on $H$, we report both the clean success rate and the Bottom10\%-SR under perturbation.
A smaller gap between these curves indicates lower sensitivity to the perturbation.
As shown in Figure~\ref{fig:action_chunk_analysis}, the gap between the clean success rate and Bottom10\%-SR widens as $H$ increases for most models. 
In contrast, the gap is small at $H=1$, where the policy can replan from a new observation at every step, limiting the persistence of erroneous actions. 
An exception is $\pi_0$-FAST, whose clean performance declines at longer execution lengths, leaving less room for further degradation.
These results support our hypothesis that action chunking amplifies vulnerability to one-step perturbations by prolonging their effects.

\section{Method}\label{sec:method}

The preceding analysis shows that shorter execution lengths improve robustness to one-step perturbations.
However, always using a short execution length does not necessarily balance robustness and task performance.
Short execution lengths can degrade performance by reducing consistency across action chunks~\citep{chi2025diffusion,black2026real} as shown in Figure~\ref{fig:action_chunk_analysis}, and increase computational cost by requiring more frequent model inference.
To address this trade-off, we propose CARE, an adaptive execution length selection method.
CARE aims to balance robustness, task performance, and computational efficiency by selecting short execution lengths only under perturbations while retaining long execution lengths for clean observations.

% \subsection{定式化}
\subsection{Formulation}
% 最初に行動列の整合性に基づいた実行長選択を定式化する．
% 時刻$t$における摂動のない観測$(v_t, s_t)$から生成される真の行動列を$\mathbf{A}^*_{t} \sim \pi_\theta(\cdot \mid v_t, s_t, l)$，実際にモデルへ入力される摂動を含む観測$(v'_t, s'_t)$において生成される行動列を$\mathbf{A}_{t} \sim \pi_\theta(\cdot \mid v'_t, s'_t, l)$とする．
% 摂動を含む観測から生成された行動列$\mathbf{A}_{t}$が真の行動列$\mathbf{A}^*_{t}$と大きく乖離している場合，行動を長く実行するほどロボットが本来の軌跡から大きく逸脱する可能性がある．
% そこで，真の行動と十分に一致する行動のみを実行し，両者の乖離が許容範囲を超える前に再推論を行うことで誤った行動の影響を抑制することができる．
% したがって，理想的な実行長$H_t^*$は行動列の先頭からの乖離が閾値以下となる最大の長さとして次式で定義される：

% We begin by formulating execution length selection based on consistency between action chunks.
% Let $\mathbf{A}^*_{t} \sim \pi_\theta(\cdot \mid v_t, s_t, l)$ be the oracle action chunk generated from the clean observation, and let $\mathbf{A}_{t} \sim \pi_\theta(\cdot \mid v'_t, s'_t, l)$ be the action chunk generated from the perturbed observation.
% If $\mathbf{A}_{t}$ differs substantially from $\mathbf{A}^*_{t}$, executing more actions may move the robot farther from the desired trajectory.
% This effect can be limited by executing only actions that remain sufficiently consistent with the clean actions and querying the policy again before the discrepancy exceeds an acceptable range.
% We therefore define the ideal execution length $H_t^*$ as the longest prefix whose discrepancy remains below a threshold:

We begin by formulating execution length selection based on consistency between action chunks.
Let $\mathbf{A}_t^* \sim \pi_\theta(\cdot \mid v_t, s_t, l)$ and $\mathbf{A}_t \sim \pi_\theta(\cdot \mid v'_t, s'_t, l)$ be action chunks generated from clean and perturbed observations at step $t$, respectively.
If $\mathbf{A}_{t}$ differs substantially from $\mathbf{A}^*_{t}$, executing more actions may move the robot farther from the desired trajectory.
We therefore define the ideal execution length $H^*$ as the longest prefix whose discrepancy from the clean action chunk does not exceed a threshold $\tau$:
\begin{equation}\label{eq:ideal_horizon}
    H^*
    =
    \max
    \left\{
        H \in \{1,\ldots,K\}
        \;\middle|\;
        D\!\left(
            \mathbf{A}_t^{1:H},
            \mathbf{A}_t^{*,1:H}
        \right)
        \leq \tau
    \right\},
\end{equation}
% ここで，$\mathbf{A}_t^{(H)}$は$\mathbf{A}_t$の先頭$H$ステップの行動からなる行動列，$D(\cdot,\cdot)$は行動列間の不一致度を表す関数，$\tau$は許容する不一致度の閾値である．
% 真の行動との乖離が小さい場合には長い実行長が選択され推論コストを低減する一方で，乖離が大きい場合には短い実行長が選択され摂動の影響を抑制する．
% where $\mathbf{A}_t^{i:j}$ denotes actions $i$ through $j$ of $\mathbf{A}_t$, inclusive, and $D(\cdot,\cdot)$ measures the discrepancy between two action chunks.
% A small discrepancy from the oracle actions yields a long execution length to reduce inference cost, whereas a large discrepancy yields a short execution length to limit the effect of the perturbation.
where $\mathbf{A}_t^{i:j}$ denotes the subsequence of actions indexed from $i$ to $j$ within $\mathbf{A}_t$, and $D(\cdot,\cdot)$ measures the discrepancy between two action chunks.
A small discrepancy from the oracle actions yields a long execution length to reduce inference cost, whereas a large discrepancy yields a short execution length to limit the effect of the perturbation.

Unfortunately, the clean observation $(v_t,s_t)$ is unavailable at inference time, so $\mathbf{A}_t^*$ cannot be computed directly.
To address this issue, we use the unexecuted portion of the previous action chunk $\mathbf{A}_{t-H_{\text{prev}}}$ as a reference for the current prediction, where $H_{\text{prev}}$ is the previous execution length.
If the environment has not changed substantially since the previous inference step, the unexecuted actions $\mathbf{A}_{t-H_{\text{prev}}}^{H_{\text{prev}}+1:K}$ should approximate the corresponding current clean actions.
For a candidate execution length $H$, we therefore compare $\mathbf{A}_t^{1:H}$ with $\mathbf{A}_{t-H_{\text{prev}}}^{H_{\text{prev}}+1:H_{\text{prev}}+H}$, which covers the same future steps.

This comparison requires at least $H$ unexecuted actions in the previous action chunk.
To satisfy this condition for every candidate execution length, CARE limits execution to $H_{\text{max}}=\lfloor K/2\rfloor$ actions per chunk.
We therefore reformulate execution length selection as
\begin{equation}\label{eq:surrogate_horizon}
H_t
=
\max
\left\{
H \in \{1,\ldots,H_{\text{max}}\}
\;\middle|\;
D\!\left(
\mathbf{A}_t^{1:H},
\mathbf{A}_{t-H_{\text{prev}}}^{H_{\text{prev}}+1:H_{\text{prev}}+H}
\right)
\leq \tau
\right\}.
\end{equation}
% Although CARE limits execution to at most half of each chunk, this is consistent with existing VLA deployment settings.
% CARE therefore requires no extension of the model's native action chunk in many cases.
% In our evaluation, the best-performing execution lengths are 10--15 for $K=50$, below $H_{\text{max}}=25$.
This execution-length cap is consistent with existing VLA deployment settings, where execution lengths are often less than half the action chunk length $K$~\citep{shukor2025smolvla}.
CARE therefore requires no extension of the model's native action chunk in many cases.

% Our consistency-based design has several advantages that simplify deployment:
% CARE requires no training, additional policy inference, or changes to the model architecture, making it easier to adopt than methods that require such modifications.
% For example, methods that select or update action chunks using multiple policy predictions incur additional inference costs~\citep{liang2026adaptive,liu2025bidirectional,so2025improving}.
% Methods that rely on learned monitors or verifiers require additional training with clean trajectories~\citep{pan2026vla,wang2026trust}.
% CARE instead reuses the unexecuted portion of the previous action chunk, which is already available at inference time, to select the execution length.

Our consistency-based design offers practical advantages for deployment.
Methods that select or update action chunks using multiple policy predictions incur additional inference costs~\citep{liang2026adaptive,liu2025bidirectional,so2025improving}, while those relying on learned monitors or verifiers require additional training~\citep{pan2026vla,wang2026trust}.
By reusing the unexecuted portion of the previous action chunk, CARE requires neither additional training nor extra policy inference for execution-length selection, and leaves the model architecture unchanged.

\subsection{Design of the Discrepancy Function}
% 本フレームワークは特定の不一致度関数に依存せず，行動間の整合性を定量化できる任意の関数を組み込むことができる．
% 本節では，この一般的な枠組みの一つの具体化を提供する。

Our framework does not depend on a particular discrepancy function and can incorporate any function that quantifies consistency between action chunks.
This section provides one concrete instantiation of this general framework.
% まず，行動ベクトルの各要素の大きさによる影響を抑えるため，
% 各相対位置$i\in\{0,\ldots,K-H_{\text{prev}}-1\}$について，時刻$t+i$に対する現在の行動$\mathbf a_{t+i}$と対応する未実行行動$\hat{\mathbf a}_{t+i}$との不一致度$r_i$を正規化ユークリッド距離を用いて以下のように計算する：
For each relative position $i\in\{0,\ldots,H_{\text{max}}-1\}$, we compute the discrepancy $r_i$ between the current action $\mathbf a_{t+i}$ and the corresponding unexecuted action $\hat{\mathbf a}_{t+i}$ using the normalized Euclidean distance to reduce sensitivity to action magnitude:
\begin{equation}\label{eq:action_inconsistency}
r_i
=
\frac{
    \left\lVert \mathbf a_{t+i}-\hat{\mathbf a}_{t+i} \right\rVert_2
}{
    \left\lVert \mathbf a_{t+i} \right\rVert_2
    +
    \left\lVert \hat{\mathbf a}_{t+i} \right\rVert_2
    + \epsilon
},
\end{equation}
% $r_i=0$は両者が一致することを表し，値が大きいほど行動の乖離が大きいことを表す．
% Larger $r_i$ values indicate greater discrepancy.
% 実行長を各時刻の不一致度のみから決定すると，摂動に起因しない予測誤差に過度に反応する可能性がある．
% 例えば，グリッパの開閉状態の反転などにより，行動列中の一時刻だけ不一致度が大きくなる場合がある．
% このような局所的な乖離は正常な観測下でも起こりうる一方で，摂動によって行動予測が大きく変化した場合には，その影響は後続する複数時刻の行動にも現れると考えられる．
% また，本手法では直前に生成された未実行行動を現在の真の行動列の代理として用いるため，予測対象が遠い未来になるほど，摂動が存在しない場合でも両者の間に自然な予測誤差が生じやすい．
% where $\epsilon>0$ is a small constant added to avoid division by zero.
% Relying only on per-step discrepancies may be overly sensitive to prediction errors unrelated to perturbations.
% For example, a one-step flip in the binary gripper action between open and closed may produce a large discrepancy.
% Such a discrepancy can occur even under clean observations, whereas a substantial change in the action prediction caused by a perturbation is expected to affect multiple subsequent actions.
% In addition, because our method uses the unexecuted actions from the preceding prediction as a surrogate for the current clean action chunk, the two chunks naturally become more likely to differ at more distant future steps, even without a perturbation.
where $\epsilon>0$ is a small constant added to avoid division by zero.
Relying solely on per-step discrepancies has two limitations.
First, the method may overreact to a large discrepancy at a single step, even without perturbations. For example, a one-step flip in the binary gripper action between open and closed may produce a large discrepancy.
Second, because we use unexecuted actions from the preceding prediction as a surrogate for the current clean action chunk, discrepancies due to natural prediction errors are more likely at more distant future steps.
% そこで，本手法では各時刻の不一致度を割引累積し，行動列全体の不一致度を次式で定義する：
We therefore define the discrepancy between action chunks as the discounted cumulative discrepancy across steps:
\begin{equation}
D
\left(
\mathbf{A}_t^{1:H},
\hat{\mathbf{A}}^{1:H}
\right)
=
\sum_{i=0}^{H-1}
\gamma^i r_i,
\end{equation}
% ここで，$\gamma\in(0,1)$は割引係数である．
% 不一致度を累積することで，単一時刻のみ生じる局所的な乖離ではなく，複数時刻にわたって継続する乖離を重視する．
% さらに，遠い未来の不一致度を割り引くことで，代理として用いる行動列の予測誤差が将来ほど大きくなる影響を抑える．
% これにより，摂動に起因しない乖離への過剰な反応を抑え，正常な観測下における実行長選択を安定化させる．
% where $\gamma\in(0,1)$ is a discount factor.
% This accumulation emphasizes discrepancies that persist across several steps. In contrast, isolated discrepancies at a single step have less influence.
% Discounting discrepancies in the distant future also reduces the influence of surrogate prediction errors, which tend to increase with the prediction horizon.
% Together, these choices prevent overreaction to discrepancies unrelated to perturbations and stabilize execution length selection under clean observations.
% We provide the pseudocode in Algorithm~\ref{alg:risk_based_action_chunking}.
where $\gamma\in(0,1)$ is a discount factor.
This accumulation emphasizes discrepancies that persist across several steps. In contrast, isolated discrepancies at a single step have less influence.
Discounting discrepancies in the distant future also reduces the influence of surrogate prediction errors.
Together, these choices prevent overreaction to discrepancies unrelated to perturbations and stabilize execution length selection under clean observations.
We provide the pseudocode in Algorithm~\ref{alg:risk_based_action_chunking}.

\section{Experiments}\label{sec:experiments}
% This section empirically evaluates our proposed method.
% We focus on the three key aspects:
% % \textbf{(i)} 提案手法によりtransient observation corruptionに対する頑健性は向上するか．
% % \textbf{(ii)} 提案手法を適用してもcleanタスク性能を維持できるか．
% % \textbf{(iii)} 提案手法による追加の計算コストはどの程度か．
% \textbf{(i)} whether the proposed method improves robustness to one-step perturbations,
% \textbf{(ii)} whether it maintains clean task performance, and
% \textbf{(iii)} how much additional computational cost it incurs.

% This section empirically evaluates our proposed method.
% We first evaluate whether CARE improves robustness to one-step observation
% perturbations while preserving clean task performance.
% We then analyze its execution length selection under clean and perturbed
% observations and quantify its computational overhead.
% Finally, we evaluate CARE on a real robot to assess its effectiveness
% in a physical environment.

We empirically evaluate our proposed method on the LIBERO~\citep{liu2023libero} and Meta-World~\citep{yu2020metaworld} benchmarks and on a real robot.
Due to space limitations, this section reports only the LIBERO and real-world results.
Implementation details and additional experiments, including the Meta-World results, are provided in Appendices~\ref{app:implementation_details} and~\ref{app:additional_experiments}.

\subsection{Setup}\label{subsec:setup}
% 提案手法のパラメータについて，割引係数を$\gamma=0.85$，最大実行長を15，初期実行長を$H_{\mathrm{init}}=1$とし，閾値は$\pi_0$，$\pi_0$-FAST，SmolVLAでは$\tau=1.5$，$\pi_{0.5}$，X-VLAでは$\tau=1.0$とする．
% 対象モデル，摂動，および評価指標については，第~\ref{sec:robustness_evaluation}章と同一の設定を用いる．

For the proposed method, we set the discount factor to $\gamma=0.85$, the maximum execution length to 15.
We set the threshold to $\tau=1.5$ for $\pi_0$, $\pi_0$-FAST, and SmolVLA, and to $\tau=1.0$ for $\pi_{0.5}$ and X-VLA.
These parameters were selected based on LIBERO-goal results without tuning on other suites.
We use the same models, perturbations, and evaluation metrics as in Section~\ref{sec:robustness_evaluation}.
\vspace{-8pt}
\paragraph{Baseline Methods.}
% 我々の知る限り、transient observation corruptionに対する頑健性を向上させることを目的として設計された手法は存在しない。
% それでも、robustnessとadaptive execution length selectionの観点から、以下の3手法と比較する．

To the best of our knowledge, no existing method is specifically designed to improve robustness to momentary observation corruptions.
% Nevertheless, we compare our method with the following three methods from the perspectives of robustness and adaptive execution length selection:
% \begin{itemize}[leftmargin=*,topsep=0pt]
%     \item \textbf{RobustVLA~\citep{guo2026on}:}
%     % 入力および出力に対する敵対的学習手法である．本実験では入力摂動のみを扱うため，入力摂動に対する敵対的学習のみを適用した．
%     This method applies adversarial training to both inputs and outputs.
%     Because our experiments consider only input perturbations, we apply only its adversarial training for input perturbations.
%     \item \textbf{VLA-Corrector~\citep{pan2026vla}:}
%     % 観測特徴の変化から現在の行動列が環境と整合しているかを判定し実行長を動的に決定する手法である．判定には事前学習した外部モジュールを利用する．Flow-matching型VLAのみに適用可能であるため，本実験では$\pi_0$-FASTを評価対象から除外する．
%     This method dynamically determines the execution length by using changes in observation features to assess whether the current action chunk is consistent with the environment.
%     It performs this assessment using a pretrained external module.
%     Because it is applicable only to flow-matching-based VLAs, we exclude $\pi_0$-FAST from its evaluation.
%     \item \textbf{AAC~\citep{liang2026adaptive}:}
%     % 同一観測に対する複数回の推論から行動分布のエントロピーを推定し，その不確実性に基づいて実行長を決定する手法である．
%     This method estimates the entropy of the action distribution from multiple inference passes on the same observation and determines the execution length based on the resulting uncertainty.
% \end{itemize}
Nevertheless, we compare our method with the three methods from the perspectives of robustness and adaptive execution length selection.
% (i) \textbf{RobustVLA~\citep{guo2026on}}: is an adversarial training method. We apply only its adversarial training for input perturbations because our experiments consider only input perturbations.
(i) \textbf{RobustVLA~\citep{guo2026on}} is an adversarial training method designed to improve robustness to persistent perturbations.
(ii) \textbf{VLA-Corrector~\citep{pan2026vla}}: dynamically determines the execution length by using changes in observation features with a pretrained external module. Since it is applicable only to flow-matching-based VLAs, we exclude $\pi_0$-FAST from its evaluation.
(iii) \textbf{AAC~\citep{liang2026adaptive}}: selects the execution length based on the entropy of the action distribution from multiple inference passes on the same observation. 
% 各ベースラインの詳細な実装についてはAppendix~\ref{app:baseline_implementation}に記載する。
We note that RobustVLA and VLA-Corrector require additional training, so the clean task performance of these methods may be affected by the training process.
% Detailed implementations of the baselines are described in Appendix~\ref{app:baseline_implementation}.

\subsection{Robustness and Clean Task Performance}\label{subsec:robustness_evaluation}

\begin{table*}[t]
    \centering
    \caption{\textbf{CARE improves robustness to one-step observation perturbations across diverse VLA models.} Clean success rate and Bottom10\%-SR (\%, $\uparrow$) across 40 tasks on LIBERO.}
    \label{tab:main_result_robustness}
    \begingroup
\scriptsize
\setlength{\tabcolsep}{3pt}
\renewcommand{\arraystretch}{1.08}
\resizebox{\textwidth}{!}{%
\begin{tabular}{llc>{\columncolor{imageinputbg}}c>{\columncolor{imageinputbg}}c>{\columncolor{imageinputbg}}c>{\columncolor{imageinputbg}}c>{\columncolor{stateinputbg}}c>{\columncolor{stateinputbg}}c>{\columncolor{stateinputbg}}c>{\columncolor{bothinputbg}}c}
\toprule
\multirow{2}{*}{Model} & \multirow{2}{*}{Method} & \multirow{2}{*}{Clean} & \multicolumn{4}{c}{\cellcolor{imageinputbg}Image input} & \multicolumn{3}{c}{\cellcolor{stateinputbg}State input} & \multicolumn{1}{c}{\cellcolor{bothinputbg}Image + state} \\
\cmidrule(lr){4-7}\cmidrule(lr){8-10}\cmidrule(lr){11-11}
 & & & Overexp. & Blackout & Blur & \shortstack{Packet\\loss} & \shortstack{Zero\\fill} & Spike & Random & \shortstack{Data loss} \\
\midrule
\multirow{5}{*}{\rotatebox[origin=c]{90}{$\pi_0$}} & Vanilla & 70.9 & 53.9 & 48.1 & 51.4 & 50.2 & 5.9 & 13.8 & 19.6 & 7.4 \\
\cdashline{2-11}
 & RobustVLA & 72.2 & 53.4 & 52.1 & 52.8 & \underline{54.1} & 10.1 & 17.2 & 21.9 & 6.9 \\
 & VLA-Corrector & 69.8 & \underline{54.2} & \textbf{53.0} & \underline{53.6} & \underline{54.1} & 9.8 & \underline{21.5} & \underline{24.2} & \underline{11.5} \\
 & AAC & 70.1 & 50.6 & 49.4 & 51.1 & 52.0 & \underline{10.2} & 10.6 & 21.0 & 9.6 \\
 & \cellcolor{black!7}\textbf{CARE (ours)} & \cellcolor{black!7}72.5 & \cellcolor{imageinputbg!93!black}\textbf{54.8} & \cellcolor{imageinputbg!93!black}\underline{52.9} & \cellcolor{imageinputbg!93!black}\textbf{54.0} & \cellcolor{imageinputbg!93!black}\textbf{56.2} & \cellcolor{stateinputbg!93!black}\textbf{51.2} & \cellcolor{stateinputbg!93!black}\textbf{47.4} & \cellcolor{stateinputbg!93!black}\textbf{50.5} & \cellcolor{bothinputbg!93!black}\textbf{53.5} \\
\midrule
\multirow{5}{*}{\rotatebox[origin=c]{90}{$\pi_0$-FAST}} & Vanilla & 89.8 & 44.4 & 45.0 & 36.7 & 75.6 & 79.7 & 72.9 & 76.1 & 49.5 \\
\cdashline{2-11}
 & RobustVLA & 91.5 & \underline{44.1} & \underline{49.9} & \underline{56.2} & \textbf{77.1} & \textbf{79.2} & \textbf{74.6} & \textbf{77.6} & \underline{55.5} \\
 & VLA-Corrector & -- & -- & -- & -- & -- & -- & -- & -- & -- \\
 & AAC & 83.0 & 42.2 & 41.6 & 33.8 & \underline{64.2} & \underline{67.5} & 64.6 & \underline{65.0} & 42.0 \\
 & \cellcolor{black!7}\textbf{CARE (ours)} & \cellcolor{black!7}78.2 & \cellcolor{imageinputbg!93!black}\textbf{62.5} & \cellcolor{imageinputbg!93!black}\textbf{63.4} & \cellcolor{imageinputbg!93!black}\textbf{64.5} & \cellcolor{imageinputbg!93!black}63.5 & \cellcolor{stateinputbg!93!black}64.4 & \cellcolor{stateinputbg!93!black}\underline{65.4} & \cellcolor{stateinputbg!93!black}63.4 & \cellcolor{bothinputbg!93!black}\textbf{64.2} \\
\midrule
\multirow{5}{*}{\rotatebox[origin=c]{90}{$\pi_{0.5}$}} & Vanilla & 97.4 & 88.0 & 45.9 & 64.9 & 91.1 & 90.1 & 91.6 & 91.0 & 47.2 \\
\cdashline{2-11}
 & RobustVLA & 97.8 & 58.1 & 53.4 & 61.3 & 88.5 & 87.9 & 87.0 & 89.4 & 53.2 \\
 & VLA-Corrector & 99.0 & \textbf{93.6} & 51.6 & \underline{70.1} & \textbf{95.6} & \textbf{96.6} & \textbf{95.9} & \textbf{97.5} & 51.4 \\
 & AAC & 96.2 & 88.5 & \underline{64.0} & 67.5 & 89.1 & 89.8 & 89.4 & 88.9 & \underline{63.6} \\
 & \cellcolor{black!7}\textbf{CARE (ours)} & \cellcolor{black!7}97.2 & \cellcolor{imageinputbg!93!black}\underline{92.1} & \cellcolor{imageinputbg!93!black}\textbf{92.0} & \cellcolor{imageinputbg!93!black}\textbf{92.2} & \cellcolor{imageinputbg!93!black}\underline{93.5} & \cellcolor{stateinputbg!93!black}\underline{93.1} & \cellcolor{stateinputbg!93!black}\underline{93.5} & \cellcolor{stateinputbg!93!black}\underline{92.1} & \cellcolor{bothinputbg!93!black}\textbf{92.4} \\
\midrule
\multirow{5}{*}{\rotatebox[origin=c]{90}{SmolVLA}} & Vanilla & 82.6 & 33.3 & 34.8 & 44.1 & 69.5 & 53.3 & 52.3 & 58.5 & 15.4 \\
\cdashline{2-11}
 & RobustVLA & 70.8 & \underline{48.4} & \underline{47.4} & \underline{48.8} & 53.8 & 32.2 & 33.2 & 44.4 & \underline{18.8} \\
 & VLA-Corrector & 86.2 & 26.4 & 32.5 & 40.4 & \textbf{75.0} & \underline{44.9} & \underline{42.9} & \textbf{63.5} & 10.1 \\
 & AAC & 76.2 & 24.9 & 25.0 & 35.1 & 61.0 & 42.1 & 32.8 & 44.2 & 18.6 \\
 & \cellcolor{black!7}\textbf{CARE (ours)} & \cellcolor{black!7}80.8 & \cellcolor{imageinputbg!93!black}\textbf{62.6} & \cellcolor{imageinputbg!93!black}\textbf{64.1} & \cellcolor{imageinputbg!93!black}\textbf{64.6} & \cellcolor{imageinputbg!93!black}\underline{67.1} & \cellcolor{stateinputbg!93!black}\textbf{58.9} & \cellcolor{stateinputbg!93!black}\textbf{60.2} & \cellcolor{stateinputbg!93!black}\underline{62.5} & \cellcolor{bothinputbg!93!black}\textbf{59.8} \\
\midrule
\multirow{5}{*}{\rotatebox[origin=c]{90}{X-VLA}} & Vanilla & 96.4 & 90.5 & 28.6 & 38.5 & 41.5 & 15.0 & 2.8 & 10.4 & 5.0 \\
\cdashline{2-11}
 & RobustVLA & 90.0 & 47.0 & 25.6 & 23.8 & 35.0 & 3.8 & 0.0 & 2.9 & 0.0 \\
 & VLA-Corrector & 93.5 & 83.8 & \underline{47.8} & 40.9 & 52.8 & 29.9 & 2.4 & 15.8 & 15.4 \\
 & AAC & 95.2 & \underline{89.0} & \textbf{57.2} & \underline{50.9} & \underline{65.4} & \underline{42.9} & \textbf{30.9} & \underline{49.5} & \underline{34.1} \\
 & \cellcolor{black!7}\textbf{CARE (ours)} & \cellcolor{black!7}96.2 & \cellcolor{imageinputbg!93!black}\textbf{91.9} & \cellcolor{imageinputbg!93!black}\textbf{57.2} & \cellcolor{imageinputbg!93!black}\textbf{65.5} & \cellcolor{imageinputbg!93!black}\textbf{72.0} & \cellcolor{stateinputbg!93!black}\textbf{86.4} & \cellcolor{stateinputbg!93!black}\underline{17.0} & \cellcolor{stateinputbg!93!black}\textbf{64.8} & \cellcolor{bothinputbg!93!black}\textbf{88.4} \\
\bottomrule
\end{tabular}}
\endgroup
\vspace{-12pt}
\end{table*}

\paragraph{Robustness to One-step Perturbations.}
% Table~\ref{tab:main_result_robustness} に、一ステップの観測摂動に対する頑健性評価の結果を示す。
% CAREは、幅広いモデルおよび摂動条件において、Vanillaモデルと比較して頑健性を改善する。
% 一部の条件、例えば $\pi_0.5$ の state input に対する摂動では、VLA-CorrectorまたはRobustVLAがCAREをわずかに上回る。
% しかし、これらはVanillaモデル自体の性能低下が比較的小さい条件であり、CAREによる改善余地も限定的である。
% 一方、Vanillaモデルで大きな性能低下が生じる摂動では、既存baselineによる性能回復は限定的であるのに対し、CAREはより安定して性能を改善する。
% 特に、RobustVLAはGaussian noiseなど学習時に想定された比較的小さな摂動に対して有効である一方、その範囲を超える大きな摂動では改善が限定的となる。
% これらの結果は、特定の摂動分布に対して頑健性を学習する既存手法と異なり、CAREがより広い強度の観測摂動に対して有効であることを示している。
% Table~\ref{tab:main_result_robustness} reports robustness to one-step perturbations.
% CARE improves robustness across a wide range of models and perturbation conditions.
% Baselines with additional training slightly outperform CARE in some conditions where Vanilla success rates drop only marginally, leaving little room for robustness gains.
% For perturbations that substantially degrade Vanilla performance, CARE improves performance more consistently, whereas existing baselines provide only limited recovery.
% In particular, RobustVLA offers limited improvements because it is effective against relatively small perturbations considered during training, such as Gaussian noise.
Table~\ref{tab:main_result_robustness} reports robustness to one-step perturbations.
CARE improves robustness particularly under perturbations that substantially degrade Vanilla performance.
In contrast, the baselines provide less consistent improvements under these conditions.
Notably, RobustVLA offers only limited gains against one-step perturbations despite its effectiveness against persistent perturbations.
Its focus on relatively small noise during training limits its effectiveness against more severe perturbations.
These results suggest that methods for improving robustness should account for severe one-step perturbations as well as persistent perturbations.

\vspace{-8pt}
\paragraph{Clean Task Performance.}
% 次に、CAREによる動的な実行長選択が、摂動のない通常条件におけるタスク性能へ与える影響を評価する。
% $\pi_0$-FASTを除くすべてのモデルにおいて、CAREによるVanillaモデルからの性能低下は最大1.8ポイントにとどまる。
% これは、CAREが摂動時の頑健性を向上させながら、多くのモデルではclean conditionにおける性能をほぼ維持できることを示している。
% For all models except $\pi_0$-FAST, the performance drop relative to Vanilla is at most 1.8 points.
% This shows that CARE largely preserves clean task performance for most models while improving robustness to perturbations.
% However, CARE reduces clean performance by 11.6 points on $\pi_0$-FAST.
% This is consistent with its sensitivity to execution length as shown in Figure~\ref{fig:action_chunk_analysis}.
% For models that perform well only within a narrow range of execution lengths, adaptive execution length selection quite often results in a trade off between clean performance and robustness.
For all models except $\pi_0$-FAST, the performance drop relative to Vanilla is at most 1.8 points.
This shows that CARE largely preserves clean task performance while improving robustness to perturbations.
Although CARE reduces clean performance by 11.6 points on $\pi_0$-FAST, this model is particularly sensitive to execution length, as shown in Figure~\ref{fig:action_chunk_analysis}.
Crucially, CARE retains substantially higher clean performance than fixed execution with $H=1$ on both $\pi_0$-FAST and X-VLA.
These results highlight that CARE improves robustness while better preserving clean performance than consistently using the shortest execution length.

% 一方、$\pi_0$-FASTでは、CAREの導入により性能が11.6ポイント低下する。
% この結果は、Figure~2で観察された、$\pi_0$-FASTが実行長の変化に対して特に敏感であり、良好な性能を示す実行長の範囲を外れると、摂動のない条件でもタスク性能が低下するという性質と整合的である。
% したがって、安定した性能を発揮できる実行長の範囲が狭いモデルでは、CAREによる動的な実行長選択がclean performanceとのトレードオフを生じさせる可能性がある。

\subsection{Execution Length Selection and Computational Cost Analysis}\label{subsec:execution_length_selection_and_computational_cost_analysis}

\begin{table}[t]
    \centering
    \caption{\textbf{CARE selects shorter execution lengths for perturbed observations.} Mean selected execution lengths for clean and perturbed observations in data-loss.}
    \label{tab:execution_length_quantitative}
    \begingroup
\scriptsize
\setlength{\tabcolsep}{3pt}
\renewcommand{\arraystretch}{1.15}
\resizebox{\textwidth}{!}{%
\begin{tabular}{@{}l*{10}{r}@{}}
\toprule
 & \multicolumn{2}{c}{$\pi_0$} & \multicolumn{2}{c}{$\pi_0$-FAST} & \multicolumn{2}{c}{$\pi_{0.5}$} & \multicolumn{2}{c}{SmolVLA} & \multicolumn{2}{c}{X-VLA} \\
\cmidrule(lr){2-3}\cmidrule(lr){4-5}\cmidrule(lr){6-7}\cmidrule(lr){8-9}\cmidrule(lr){10-11}
Method & Clean & Perturbed & Clean & Perturbed & Clean & Perturbed & Clean & Perturbed & Clean & Perturbed \\
\midrule
VLA-Corrector & 9.20 & 9.16 & -- & -- & 9.51 & 9.71 & 8.60 & 8.12 & 8.41 & 8.29 \\
AAC & 18.97 & 14.81 & 8.92 & 9.28 & 15.51 & 26.05 & 19.81 & 13.52 & 10.78 & 9.11 \\
\rowcolor[gray]{0.93}
\textbf{CARE (ours)} & 8.98 & 2.22 & 6.56 & 1.29 & 10.53 & 1.39 & 10.65 & 2.45 & 11.45 & 1.04 \\
\bottomrule
\end{tabular}
}
\endgroup
\vspace{-8pt}
\end{table}

\begin{figure}[t]
    \centering
    \begin{minipage}[t]{0.50\textwidth}
        \vspace{0pt}
        \centering
        \includegraphics[width=\linewidth]{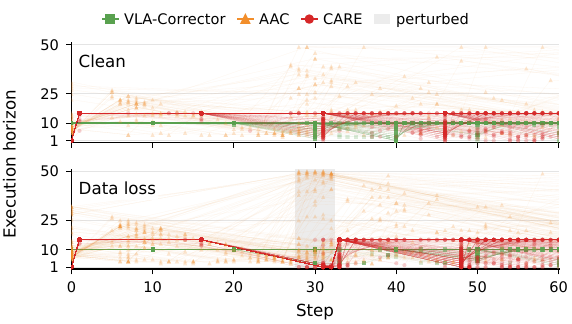}
        \vspace{-7mm}
        \caption{\textbf{Execution length selection for $\pi_{0.5}$.}
        The shaded band indicates the range of intervention steps; each trial is perturbed at only one step.}
        \label{fig:execution_length_selection}
    \end{minipage}\hfill
    \begin{minipage}[t]{0.48\textwidth}
        \vspace{0pt}
        \centering
        \includegraphics[width=\linewidth]{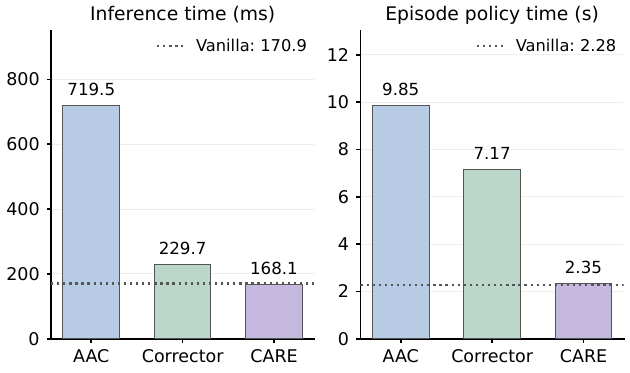}
        \vspace{-7mm}
        \caption{\textbf{Computational cost comparison.} Inference latency and cumulative policy time per episode.}
        \label{fig:computational_cost}
    \end{minipage}
    \vspace{-12pt}
\end{figure}

\paragraph{Execution Length Selection.}
Table~\ref{tab:execution_length_quantitative} reports the mean selected execution lengths under data loss.
CARE maintains long execution lengths for clean observations and substantially shortens them for perturbed observations, whereas the baselines show less consistent changes.
% Figure~\ref{fig:execution_length_selection} illustrates this behavior for $\pi_{0.5}$ using trials in which the perturbation occurs at a step between 28 and 32, alongside clean trials.
Figure~\ref{fig:execution_length_selection} illustrates this behavior for $\pi_{0.5}$ by comparing clean trials with perturbed trials at a single step between steps 28 and 32.
Execution lengths shorten sharply at the perturbed step indicated by the shaded band.
% Under clean conditions, CARE mainly selects long execution lengths.
% Under perturbed conditions, the selected lengths cluster sharply at small values at the perturbed steps.
These results provide both quantitative and qualitative evidence that CARE selects shorter execution lengths for perturbed observations.

\paragraph{Computational Cost.}
% Figure4に、clean runにおける全てのモデルの平均の1回の推論の処理時間と、1episodeの平均処理時間を示す。
% 提案手法の処理時間は，適用前と比べて 1 ms 未満の増加にとどまり，実行長選択の追加コストが小さいことが確認できる．これは，提案手法が既に生成された action chunk のみを利用して実行長を決定するため，追加の VLA 推論を必要としないことによる．一方，baselinesでは、2倍以上のコスト増加が発生する。AAC ではエントロピー推定のために複数回の VLA 推論を実行する必要があり，VLA-Corrector では外部モジュールによる追加推論が必要となるため，計算時間が増加したと考えられる．
% Figure~\ref{fig:computational_cost} compares clean-run inference latency and cumulative policy time per successful episode.
Figure~\ref{fig:computational_cost} shows computational costs under clean observations: the average time to generate an action chunk and select its execution length, and the average total computation time per successful episode.
We evaluate 40 tasks with five initial conditions each, using one environment and the same H100 GPU across methods for each task group.
The plots average $\pi_0$, $\pi_{0.5}$, SmolVLA, and X-VLA equally; $\pi_0$-FAST is excluded because VLA-Corrector does not support it.
CARE achieves inference latency and cumulative policy time per successful episode comparable to Vanilla, with only a slight increase.
This is because CARE selects the execution length using only the already generated action chunk, without requiring additional VLA inference.
In contrast, AAC requires multiple candidate action chunks to estimate entropy, and VLA-Corrector requires additional inference with an external module, leading to increased computation time.

% \begin{table}[htbp]
%     \centering
%     \caption{\small Decision latency (ms, $\downarrow$); parentheses: ratio to Vanilla.}
%     \label{tab:proposed_method_computational_cost}
%     \scriptsize
%     \setlength{\tabcolsep}{2pt}
%     \setlength{\dashlinedash}{1.5pt}
%     \setlength{\dashlinegap}{1.5pt}
%     \renewcommand{\arraystretch}{1.08}
%     % Values and ratios reproduced from the CSS2026 computational-cost table.
%     \begin{tabular}{@{}p{0.16\linewidth}*{5}{>{\centering\arraybackslash}p{\dimexpr0.168\linewidth-2\tabcolsep\relax}}@{}}
%         \toprule
%         Method & $\pi_0$ & $\pi_0$-FAST & $\pi_{0.5}$ & SmolVLA & X-VLA \\
%         \midrule
%         Vanilla & 194.0 & 2460.4 & 212.3 & 151.6 & 117.5 \\
%         \hdashline
%         \noalign{\vskip 1pt}
%         VLA-Corrector & 576.0 (2.969$\times$) & -- & 607.4 (2.861$\times$) & 313.3 (2.066$\times$) & 302.1 (2.571$\times$) \\
%         AAC & 711.1 (3.666$\times$) & 3192.7 (1.298$\times$) & 775.2 (3.652$\times$) & 185.1 (1.221$\times$) & 1181.7 (10.055$\times$) \\
%         \rowcolor[gray]{0.93}
%         \textbf{CARE (ours)} & \textbf{194.6} (1.003$\times$) & \textbf{2461.2} (1.000$\times$) & \textbf{212.8} (1.003$\times$) & \textbf{152.1} (1.003$\times$) & \textbf{117.8} (1.003$\times$) \\
%         \bottomrule
%     \end{tabular}
% \end{table}

\subsection{Real-World Robot Experiments}\label{subsec:real_world_robot_experiments}
% \paragraph{Setup.}
% - Trossen Stationary AIを使用したことを説明する（robot armと4つのカメラについて）。
% - pick-and-placeタスクを行ったことを説明する。right armでblockをつかんで、left armに渡して、left armでblockをplaceするタスクを行ったことを説明する。1episodeが30秒で構成され、学習されたモデルはおよそ20秒でタスクを完了することを説明する。
% - 学習について説明する。pi05をbaseにして、100episodeのdemoを用いて、40000stepをloraで学習した。rank / alphaは32 / 32で、batch size = 16, adapWを使用。学習率は2.5e-5.学習されたモデルは、clean conditionで100\%の成功率を達成したことを説明する。  実効長は25にする。 
% - 出力にguard railをつけたことを説明。最初はつけてなかったけど、摂動を入れると非常に危険な振る舞いをしたことから、出力を[]に制約する。

% \paragraph{Perturbation Design.}
% - 摂動には、data loss, random state, blurを使用する。実際にロボットを動かしたりすることはcontrolledな環境ではできないので、入力に摂動を人工的に入れる。
% - SImulationでの実験と同様に、全ての推論に摂動を入れるのは高コストなので、私たちは1episodeの振る舞いを分割して、それぞれに摂動を入れる。
% - 具体的には、私たちはclean runにおけるロボットアームの動きから、1episodeにおけるモデルの振る舞いを4phaseに分割する (Figure3)。(i) 物体の探索: right armが物体をカメラで捉える段階. (ii) 物体の把持; right armが、ブロックを掴み持ち上げる段階。 (iii) 持ち替え: right armが持っているブロックをleft armに渡す段階。(iv) 物体の配置: left armがブロックを配置する段階。物体の各phaseは100stepで構成される。
% - 私たちは、各phaseにおいて、推論を行うstepをランダムに選択して摂動を挿入する。私たちはこれによって、WSRを計算する。
\vspace{-4pt}
\paragraph{Setup.}
We deploy the VLA on two 6-DoF WidowX AI follower arms with four Intel RealSense D405 cameras.
We evaluate the VLA on a pick-and-place task in which the right arm grasps a red block and passes it to the left arm, which then places it at a designated location.
% Each episode lasts 40 seconds, although the trained policy typically completes the task in approximately 20 seconds, including inference time.
We fine-tune $\pi_{0.5}$ on 100 demonstration episodes and set the execution length of $H=25$.

\begin{figure}[t]
    \centering
    \includegraphics[width=\linewidth]{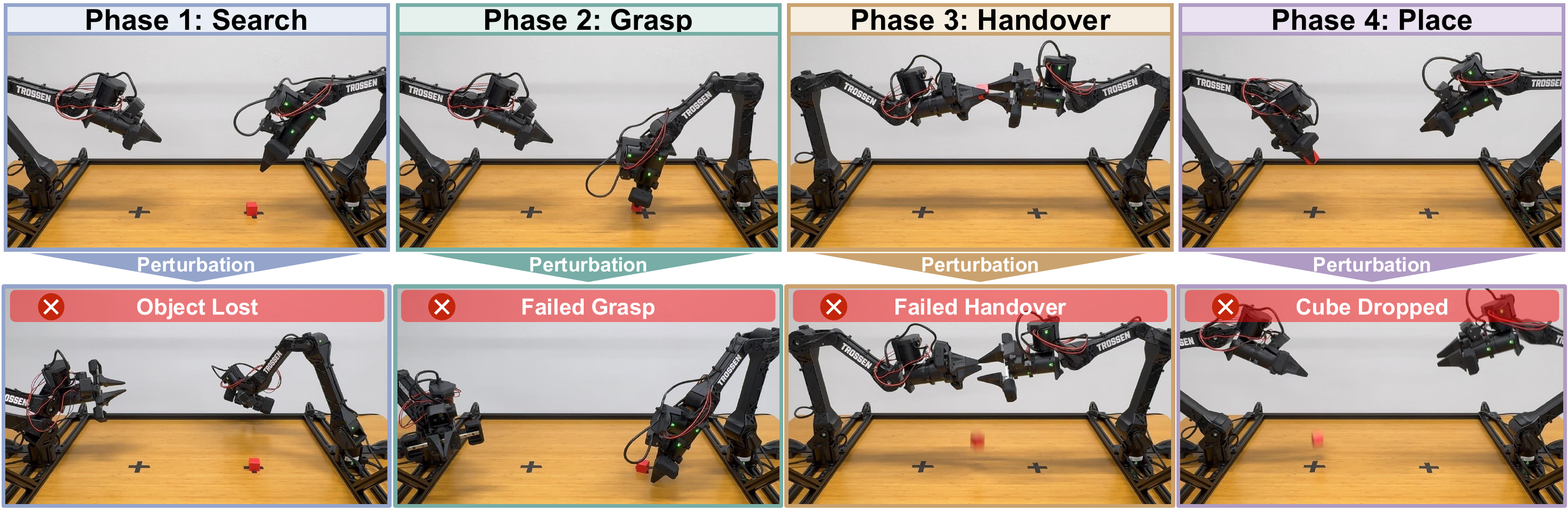}
    \vspace{-7mm}
    \caption{\textbf{Typical failure examples in real-world robot experiments.}
    The top row shows the four task phases before perturbation, and the bottom row shows typical failures following a one-step perturbation in each phase.}
    \label{fig:real_world_failure_examples}
    \vspace{-12pt}
\end{figure}

\vspace{-8pt}
\paragraph{Perturbation Protocol.}
We evaluate three perturbation types: data loss, random state, and blur.
To ensure controlled evaluation conditions, we apply synthetic perturbations rather than physically disturbing the robot or its environment.
% Our preceding experiments show that the effect of perturbations depends on the robot's state.
% To assess this dependence, we divide the task into four phases based on unperturbed execution, as shown in Figure~\ref{fig:real_world_failure_examples}:
To examine the effect of perturbation timing, we divide the task into four phases based on unperturbed execution, as shown in Figure~\ref{fig:real_world_failure_examples}:
(i) search: the right arm brings the block into camera view,
(ii) grasping: it grasps and lifts the block,
(iii) handover: it transfers the block to the left arm, and
(iv) placement: the left arm places the block at the designated location.
Each phase consists of 100 control steps.
For each target phase, we apply the perturbation at one randomly selected inference step within that phase.
See Appendix~\ref{app:real_world_experiment_details} for details.

% \paragraph{Results.}
% As shown in Table~\ref{tab:main_result_robustness}, vulnerability to one-step perturbations is a significant challenge even in real-world settings. 
% vanilla modelは評価した全てのperturbationにおいて、成功率が大幅に減少していることが確認できる。
% 一方で、CAREはreal-worldの全てのperturbationにおいて、Vanillaモデルと比較して成功率を大幅に改善していることが確認できる。これは、simulationでの結果と同様に、実効長の動的選択が、摂動に対する頑健性を改善することを示している。
% また、我々は摂動によって生じるfailure modeをFigure~\ref{fig:real_world_failure_examples}に示す。
% 我々は、雪道の種類によって生じるfailure modeが異なることを確認した。例えば、data lossとrandom stateでは、摂動が挿入された瞬間、ロボット自体が大きく動き出し、demonstrationには存在しない状態に遷移してしまうことがある。この状態になると、ロボットは物体を見失い、動かなくなってしまうことがある。一方で、phase3,4などの既に物体を把持している状態では、摂動が挿入されても、カメラがブロックをとらえているため、そこから復帰することがあった。そのため、このような摂動では、phase1,2での失敗が多く、phase3,4では比較的成功率が高いことが確認できた。
% 一方で、blurでは、摂動を挿入してもロボットが大きく動き出すことはなかった。これは、demo中でも、カメラの映像がblurしている状態が存在するため、blurに対しては、ロボットが大きく動き出すことはなかったと考えられる。しかし、blurでは、物体を把持しているときに、摂動が挿入されると、物体を落とすといったfailure modeが頻発した。そのため、他の雪道とは逆に、blurでは、phase1,2での失敗は少なく、phase3,4での失敗が多いことが確認できた。

\begin{figure}[!t]
    \centering
    \includegraphics[width=\linewidth]{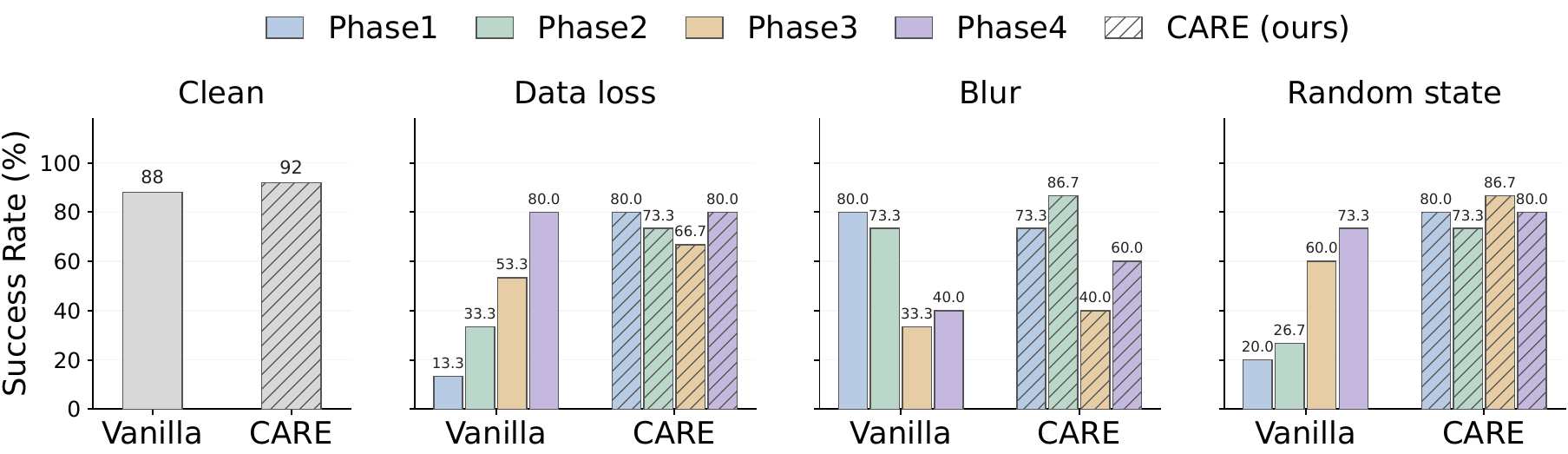}
    \vspace{-7mm}
    \caption{\textbf{Real-world evaluation results.}
    Success rates are averaged over 25 for clean conditions and 15 episodes for perturbed conditions.}
    \label{fig:real_world_success_by_phase}
    \vspace{-6pt}
\end{figure}

% \vspace{-8pt}
% \paragraph{Results.}
% Figure~\ref{fig:real_world_success_by_phase} reveals real-world vulnerabilities to one-step perturbations. Every perturbation reduces the success rate to 33.3\% or lower in at least one phase. As in simulation, performance depends on perturbation type and timing. Data loss and random state cause larger declines in early phases, whereas Blur has a greater impact in late phases.
% We illustrate these failure modes in Figure~\ref{fig:real_world_failure_examples}. Before grasp (Phases 1 and 2), data loss and random state often caused large robot movements followed by a halt. After grasp, the robot sometimes recovered, which suggests that a block within the camera’s view may help prevent OOD states and enable recovery. Blur caused no large movements but often caused the gripper to release the block. This may explain the larger performance decline in late phases, when the robot often held the block.
% CARE improved real-world robustness, but its gains against Blur were limited. The gripper could release the block within a single step, so a shorter execution length could not prevent the drop. Thus, execution length adjustments alone may not address failures caused by momentary errors.

\vspace{-8pt}
\paragraph{Results.}
Figure~\ref{fig:real_world_success_by_phase} reports real-world success rates under one-step perturbations.
All three perturbations reduce Vanilla's success rate to 33.3\% or lower in at least one phase.
As in simulation, performance depends on perturbation type and timing.
Figure~\ref{fig:real_world_failure_examples} illustrates these failure modes.
Data loss and random state cause larger performance declines during search and grasping, often inducing large robot movements followed by a halt.
After grasping, however, the robot sometimes recovers while the block remains within camera view.
This suggests that target visibility may facilitate recovery.
Blur causes larger performance declines during the handover and placement phases.
It does not induce large robot movements but often causes the gripper to release the block.
This failure mode may explain its greater impact in later phases, when the robot is holding the block.
CARE improves real-world robustness, although its gains against blur are limited.
The gripper can release the block within a single control step, so even the shortest execution length may not prevent the drop.
These results suggest that execution length adjustment alone may be insufficient to prevent failures caused by a single erroneous action.

\section{Conclusion}\label{sec:conclusion}
% Original draft.
% 本研究では、one-step perturbationに対するVLAの顔牽制評価を行った。
% 実験の結果、摂動によってVLAの性能が大きく低下することを明らかにした。
% また，action chunking によって誤った行動列が複数ステップにわたり実行されることが脆弱性の一因であることを示した．
% この知見に基づき，本研究では予測された行動列の整合性に応じて実行長を動的に調整する手法であるCAREを提案した.
% Emprical evaluation shows that CAREが多くのモデルで通常時のタスク性能を概ね維持しながら単一時刻の入力摂動に対する頑健性を向上できることを確認した．
% Our findings open several promising directions for future research. Future work could explore temporally localized perturbations in broader security settings, including backdoor attacks with short-lived triggers, adversarial attacks targeting critical moments, and defenses against such transient interventions. Our results also motivate new evaluation protocols and benchmarks that consider not only perturbation magnitude and form, but also timing and duration. This temporal perspective may reveal failure modes overlooked by conventional persistent-perturbation evaluations and enable more comprehensive robustness assessment of sequential decision-making systems.
% Revised Japanese draft.
% 本研究では、単一の推論時刻における観測破損だけでも、VLAのタスク成功率が大幅に低下し得ることを示した。
% また、長い実行長によって摂動の影響が後続のステップまで続き得ることを示した。
% この分析に基づいて提案したCAREは、行動チャンク間の整合性に応じて実行長を調整し、多くのモデルで通常時の性能を概ね維持しつつ頑健性を改善した。
We showed that observation corruption at a single step can substantially reduce VLA task success rates and that long execution lengths can prolong its effects into subsequent steps.
Our proposed method, CARE, adapts the execution length based on consistency between action chunks, improves robustness, and largely preserves clean performance for most models.
% Our evaluation focuses on finite-horizon tasks, and CARE provides smaller robustness gains against multi-step perturbations. Further discussion appears in Appendix~\ref{app:discussion}.
While this work focuses on naturally occurring corruptions, our findings open several promising directions for future research. Future work could explore temporally localized perturbations in broader security settings, including backdoor attacks with short-lived triggers, adversarial attacks targeting critical moments, and defenses against such momentary interventions. This temporal perspective may reveal failure modes overlooked by existing evaluations.

% \paragraph{Limitations.}
% % Original draft.
% % 本提案手法は、何ステップも連続する摂動には対処できない。
% % 対処するには毎ステップの監視がいるけど、これは本質的に追加の計算コストを必要とし、実用性が低下する。
% % また、一度崩壊したVLAを復帰させることはできない。
% % Future workとして、摂動の影響を最小化するのとは別に、摂動によって崩壊してからの復帰力を高めるための学習手法の開発が求められる。

% % CAREはmomentary observation corruptionを対象としており、再推論時にも観測が破損している場合には、実行長の短縮だけでは正常な入力に基づく制御を再開できない。
% % こうした摂動に対する解決策として、各ステップでの監視があるが、こうした手法には追加の計算コストが伴う。
% % また、CAREは誤った行動の実行を抑制する手法であり、すでに大きく逸脱した状態からの回復を保証するものではない。
% % 摂動の影響を最小化するのとは別に、摂動によって崩壊してからの復帰力を高めるための学習手法との組み合わせが今後の課題である。
% If observations remain corrupted at subsequent inference steps, CARE cannot restore control based on clean inputs by shortening the execution length alone.
% Monitoring observations at every step offers a potential way to address such perturbations but incur additional computational cost.
% Moreover, CARE limits the execution of erroneous actions but does not guarantee recovery once the robot has substantially deviated from its intended trajectory.
% Future work should combine CARE with methods that train policies to recover from perturbation-induced failures.

\subsection*{AI use statement}

% 私たちは、生成AIを、実験コードの実装、および論文の文章校正に使用しました。
% Research ideaを出すことについてはAIは一切使用していません。
% 生成されたものについては、全て人間による監査が入っています。
% 例えば、AIにより執筆された文章は、その後人間により確認、または修正されています。
% LLMで生成されたコードは著者が正確性を確認しています。

We used generative AI tools to assist with the implementation of experimental
code and the writing and language editing of this manuscript. We did not use
generative AI to formulate the research ideas underlying this work. All
AI-assisted outputs were reviewed by the authors: AI-generated or AI-edited text
was checked and, where necessary, revised, and LLM-generated code was inspected
and tested for correctness. The authors take full responsibility for the final
content of this work.

\subsection*{Ethics statement}
We use publicly available datasets and benchmarks for simulation experiments and collect robot demonstrations for real-world experiments. Our study involves neither human-subject experiments nor personal data.
Although our robustness evaluation
exposes failure modes that could potentially be misused, we also propose and
evaluate a countermeasure, with the goal of improving the robustness and safety
of VLAs. We have no conflicts of interest or sponsorship to disclose. We
conducted this work in accordance with applicable ethical guidelines and
research-integrity standards.

\subsection*{Reproducibility statement}
The experimental settings and hyperparameters are reported in
Sections~\ref{subsec:experimental_setup} and~\ref{subsec:setup}. Additional
implementation details for our method and all baselines are provided in
Appendix~\ref{app:implementation_details}. All benchmarks used in our
experiments are publicly available to the research community.

\bibliographystyle{iclr2027_conference}
\bibliography{ref}

\appendix
% !TeX root = preprint.tex

\section{Related Work}\label{app:related_work}

\subsection{Vision-Language-Action Models}\label{subsec:related_work_vla}
% VLAは大規模な画像言語データセットで学習された Vision-Language-Model (VLM) の視覚・言語理解能力をロボットの行動生成へと拡張したモデルである~\citep{brohan2023rt,zitkovich2023rt,mees2024octo,kim2024openvla,black2024pi_0,black2025pi,pertsch2025fast,shukor2025smolvla,wen2025dexvla,zheng2026xvla,liu2025rdt}。
% このとき、VLMの出力をロボットの制御信号に変換する専用モジュールであるaction expertが組み合わされることが多い。最初期のVLAでは、言語と同様に、行動を離散的なトークン列として自己回帰的に生成する手法が提案されている~\citep{brohan2023rt,zitkovich2023rt,kim2024openvla}。
% 近年では、より高精度な行動を生成するために、diffusion model~\citep{mees2024octo}やflow-mathching~\citep{black2024pi_0,black2025pi}を用いた連続的な行動生成手法が提案されている。
% 既存研究では、主に正常な観測下におけるタスク性能が評価されてきた一方で、本研究では推論中の入力に対するtransient perturbationに対する頑健性に焦点を当てる。

VLA models extend the visual and linguistic understanding capabilities of Vision-Language Models (VLMs), trained on large-scale vision-language datasets, to robotic action generation~\citep{brohan2023rt,zitkovich2023rt,mees2024octo,kim2024openvla,black2024pi_0,black2025pi,pertsch2025fast,shukor2025smolvla,wen2025dexvla,zheng2026xvla,liu2025rdt}.
They are often paired with an action expert, a specialized module that converts VLM outputs into robot control signals. 
Early VLAs generated actions autoregressively as sequences of discrete tokens, in the same manner as language~\citep{brohan2023rt,zitkovich2023rt,kim2024openvla}.
More recently, continuous action generation methods based on diffusion models~\citep{mees2024octo} and flow-matching~\citep{black2024pi_0,black2025pi} have been proposed to generate more precise actions.
While prior work has primarily evaluated task performance under clean observations, we focus on robustness to input perturbations during inference.

\subsection{Robustness to Input Perturbations}\label{subsec:related_work_robustness}
% VLAモデルは，物理世界で動作するロボットの行動を生成するため，誤った行動はタスク失敗だけでなく周囲の環境や人への危害につながる可能性がある．
% そのため，VLAを安全に実環境へ導入するために、多くの研究がVLAに対する安全性について研究している~\citep{zhang2025badrobot,robey2025jailbreaking,jones2025adversarial,li2026vision}。 

% Because VLA models generate actions for robots operating in the physical world, incorrect actions may not only cause task failure but also harm the surrounding environment or people.
% Consequently, many studies have investigated VLA safety to support their safe deployment in real-world environments~\citep{zhang2025badrobot,robey2025jailbreaking,jones2025adversarial,li2026vision}.

% 特に推論時には，攻撃者が画像や状態入力へ直接介入できることから，入力摂動に対する頑健性が重要な課題となる．
% 近年では画像入力や状態入力，あるいはその両方に対する多様な摂動を用いてVLAの頑健性を評価する研究が行われており，現在のVLAはこれらの摂動によってタスク成功率が大きく低下することが報告されている~\citep{guo2026on,wang2025vlatest,hancock2025run,wang2025exploring,lu2026phantom,xie2026strong,zhang2025robustvla,xu2025model}
% しかし，これらの研究では，タスク実行中のすべての推論時刻において継続的に入力へ摂動が与えられることを仮定している．
% 物理世界では、継続的な摂動よりも、通信障害や接触などによって発生する一時的な摂動の方が現実的である。

Because VLA models act in the physical world, their safety is critical, as incorrect actions can cause task failure, damage nearby objects, or harm people.
A growing body of work has therefore investigated VLA safety to support safe deployment in real-world environments~\citep{zhang2025badrobot,robey2025jailbreaking,jones2025adversarial,li2026vision}.

One focus of this research is robustness to input perturbations during inference.
At deployment, VLA models receive observations from environments they do not control, where external factors can corrupt image or state inputs.
Understanding how VLA models behave under such corruption is therefore important for assessing their safety.
Recent studies have evaluated VLA robustness using various perturbations to image inputs, state inputs, or both, and have reported substantial drops in the task success rate under these perturbations~\citep{guo2026on,wang2025vlatest,hancock2025run,wang2025exploring,lu2026phantom,xie2026strong,zhang2025robustvla,xu2025model,li2025cronusvla}.
However, these studies mainly assume persistent perturbations, which are continuously applied throughout the entire task execution.
% However, these studies assume persistent perturbations where the inputs are continuously perturbed at every inference step throughout task execution.
% In the physical world, transient perturbations caused by events such as communication failures or contact are more realistic than continuous perturbations.
% We note that attacks on langauage instuructions, such as jailbreaks~\citep{robey2025jailbreaking,zhang2025badrobot}, are out-of-scope because the language instruction is provided once at the beginning of an episode and remains fixed during subsequent inference, making such attacks fundamentally different from transient perturbations considered in this work.

% 本研究では，VLA の時系列的な意思決定に着目し，単一の時刻にのみ介入する摂動に対する頑健性を評価し，VLA の新たな脆弱性を明らかにする．
% We note that 本研究ではjailbreakなどの言語指示への攻撃~\citep{robey2025jailbreaking,zhang2025badrobot}は対象外である。なぜなら多くのVLAでは言語指示はエピソード開始時に一度与えられ，以降は固定されたまま推論が行われるため，本研究が対象とする推論途中の入力介入とは性質が異なるためである。

In this work, we evaluate the robustness against one-step input perturbations that intervene at only a single inference step and reveal a new vulnerability of VLAs.
Furthermore, we analyze this vulnerability and show that action chunking is one factor that prolongs the effects of a perturbation into subsequent steps.

\subsection{Action Chunking}\label{subsec:related_work_action_chunking}
% Action chunking~\citep{zhao2023learning,chi2025diffusion}は，1回の推論で複数ステップ分の将来の行動列を生成し，そのうち事前に定められた実行長分の行動を順次実行する推論機構である．
% モデルの推論回数を削減することにより計算コストを抑えながら短い制御周期でのロボット制御を可能にする一方で，一度決定した行動を変更することができないため，急速な行動変化に対する反応性が低下するというトレードオフが存在する．

Action chunking~\citep{zhao2023learning,chi2025diffusion} is an inference mechanism that predicts a sequence of future actions spanning multiple steps in a single inference pass and then sequentially executes a predetermined number of those actions.
By reducing the number of model inference calls, action chunking enables robot control at short control intervals while limiting computational cost. However, because previously determined actions cannot be changed, it introduces a trade-off in which responsiveness to rapid changes in motion is reduced.

% このトレードオフを改善するために，観測に合わせて実行長を動的に変化させる手法が提案されている~\citep{wang2026trust,liang2026adaptive,pan2026vla,wang2026vla,jing2025mixture}．
% しかし，これらの手法は観測に加えられる摂動を考慮しておらず，モデルの頑健性への寄与は限定的である．
% 本研究では，摂動に対する頑健性向上を目的として，予測されたaction chunkから得られる情報のみを用いて実行長を動的に調整する新たな手法を提案する．
% 提案手法は，追加のモデル推論や学習を必要とせず，特定の行動生成方式に依存しないため，action chunkを出力する幅広いVLAへ適用できる．

Several methods have been proposed for adaptive execution length selection based on the observation to mitigate this trade-off~\citep{wang2026trust,liang2026adaptive,pan2026vla,wang2026vla,jing2025mixture}.
However, these methods do not consider perturbations to the observations, limiting their contribution to model robustness.
To improve robustness to perturbations, we propose CARE, an adaptive execution length selection method that uses only information obtained from the predicted action chunk.
Our method requires neither additional model inference nor training and is independent of any particular action generation method. It can therefore be applied to a wide range of VLAs that output action chunks.

\subsection{Consistency-Based Methods}\label{subsec:related_work_consistency}
Several methods exploit consistency across action chunks.
Sentinel~\citep{pmlr-v270-agia25a} detects erratic failures by measuring the consistency of action distributions over temporally overlapping portions of successive predictions.
BID~\citep{liu2025bidirectional} generates multiple candidate chunks at each control step and selects among them using criteria that include consistency with the preceding prediction.
However, neither method is designed to adaptively select the execution length.
SGAC~\citep{so2025improving} predicts a new action chunk at every step and compares its first action with the next queued action to decide whether to retain or update the existing plan.
Although this approach adaptively updates the plan, it requires policy inference at every step, incurring greater inference demand than executing multiple actions between policy calls.

Our proposed method uses consistency across action chunks to improve robustness to one-step perturbations by adaptively selecting the execution length.
This selection requires no additional policy inference and avoids the need to query the policy at every step.

\section{Discussion and Limitations}\label{app:discussion}

\vspace{-4pt}
\paragraph{Evaluation on infinite-horizon tasks.}
% 本研究の評価はfinite-horizonタスクに限定されている。Infinite-horizonタスクでは、one-step perturbationによって一時的に不適切な行動が生じても、その後の時間を使って修正できるため、摂動の影響が一時的な区間にとどまる可能性がある。しかし、物体を落として破損させるなどの不可逆的な失敗は、その後の行動によって修正できず、infinite-horizonタスクにおいても依然として脅威となる。したがって、こうした不可逆的な失敗につながる局面を考慮したmomentary perturbationsの設計と評価は、重要な今後の課題である。
Our evaluation is limited to finite-horizon tasks. In infinite-horizon tasks, an agent may have time to recover from misbehavior caused by a one-step perturbation. It potentially confines its effects to a brief interval. However, irreversible failures, such as dropping and breaking an object, cannot be corrected through subsequent actions and therefore remain a threat even in infinite-horizon tasks. Designing and evaluating momentary corrupting in infinite-horizon settings is thus an important direction for future work.

\vspace{-8pt}
\paragraph{Robustness against multi-step perturbations.}
% CAREはone-step perturbationsに対して有効性を示す一方、数十ステップにわたる摂動に対する効果には限界がある。Appendix~\ref{app:multistep_perturbations}に示すように、摂動が適用されるステップ数が増えるほど、CAREによる頑健性の改善は限定的になる。これは、どのような実行長を選択しても摂動の影響が複数ステップにわたって続くため、実行長の選択だけではその影響を十分に抑えられないためと考えられる。こうした摂動への対応には、失敗軌道からの回復を学習する手
Although CARE is effective against one-step perturbations, its effectiveness is limited against perturbations lasting 5--10 steps.
As shown in Appendix~\ref{app:multistep_perturbations}, CARE provides smaller robustness gains as the number of perturbed steps increases. A possible explanation is that the perturbation continues to affect multiple steps regardless of the chosen execution length, limiting the extent to which execution-length selection alone can mitigate its effects. Addressing such perturbations may therefore require combining CARE with methods that learn to recover from failure trajectories.

\section{Implementation Details}\label{app:implementation_details}

\subsection{Computational Environment}
Experiments were executed on a HPC cluster. 
We use single NVIDIA H100. The system ran NVIDIA driver 575.57.08 with CUDA 12.9.

\subsection{Perturbation Implementation}\label{app:perturbation_implementation}
We apply each perturbation to the observation after environment-level
preprocessing and before model inference. Image perturbations are applied to every input image, whereas state perturbations are applied to the state observations. Unless otherwise noted, a perturbation is active for one step at the specified inference step. 
We adapt the image-corruption designs from \citet{zeng2024benchmarking}.

\paragraph{Blackout.}
We replace every value in each camera image with zero while preserving the
original shape and data type. 
Thus, a normalized floating-point image is mapped to an all-zero tensor, and an 8-bit image is mapped to an all-black image. The state observation is left unchanged.

\paragraph{Overexposure.}
For every pixel value $x$, we compute
$x'=(1-\alpha)x+\alpha x_{\max}$ with $\alpha=0.90$, where $x_{\max}=1$ for
floating-point images represented in $[0,1]$ and $x_{\max}=255$ otherwise. 
We clip the result to the corresponding valid range and cast it back to the original data type. This transformation is applied independently to every camera image and leaves the state observation unchanged.

\paragraph{Motion blur.}
We first convert each camera image to an 8-bit representation and apply the
OpenCV~\citep{opencv_library} convolution with a normalized line kernel of length $41$. 
The blur direction is sampled uniformly from $[-45^{\circ},45^{\circ}]$ for each image, and the kernel is rotated to the sampled angle before convolution. The result is clipped to $[0,255]$ and then restored to the original data type and numerical range; the state observation is not modified.

\paragraph{Packet loss.}
We model packet loss by combining stale rectangular regions with local block artifacts. For each camera stream, we retain the most recent clean image and copy 20 randomly positioned horizontal rectangles from that image into the current image. Before clipping at image boundaries, each rectangle has a height between 10 and 24 pixels and a width between 150 and 199 pixels. We then sample $2\times700=1{,}400$ additional locations and replace each local block, whose height and width are determined by the sampled coordinates modulo 10, with the pixel value at its upper-left corner. The state observation is left unchanged.

\paragraph{Zero fill.}
We replace all components of a state observation with zero while preserving the shape and data type of the state vector. All camera observations remain unchanged.

\paragraph{State spike.}
We replace the first three components of a state observation, which represent the end-effector Cartesian position, with their upper bounds. When finite bounds cannot be obtained from the environment observation space, we use the fallback upper bound $1$ for each of these components. The remaining state components and all camera observations are left unchanged.

\paragraph{Random state.}
We replace every state component independently with a signed outlier. For state dimension $j$ with absolute upper bound $b_j$, we sample its magnitude uniformly from $[b_j,2b_j]$ and its sign uniformly from $\{-1,+1\}$. If finite bounds cannot be obtained from the environment observation space, we use $(1,1,1,\pi,\pi,\pi,1,1)$ as the fallback bounds for the eight-dimensional state vector. Camera observations are not modified.

\paragraph{Data loss.}
We apply Blackout and Zero fill simultaneously: every camera image and every component of a state observation are replaced with zero, with their original shapes and data types preserved.

\subsection{Baseline Implementation}\label{app:baseline_implementation}

% 私たちは、ベースラインとして3つの手法を選択した。我々の知る限り、momentary observation corruptionに対して設計された手法は存在しなかった。
% そのため、私たちはRobustVLA、VLA-Corrector、AACを選択した。本セクションでは、それぞれの手法の実装について記載する。
In this work, we consider three baselines: RobustVLA, VLA-Corrector, and AAC. To the best of our knowledge, no existing method is specifically designed for momentary observation corruption. We therefore select these methods as representative approaches to robustness and adaptive execution length selection. This section describes how each baseline is implemented in our experiments.

\paragraph{RobustVLA.}
This method improves robustness by augmenting the native policy objective with adversarial objectives for perturbed inputs and outputs. Since we consider only input perturbations, we retain only its input-robustness objective and optimize the sum of the native clean loss and the adversarial image loss. 
At each update, UCB selects one of eight visual augmentation choices, after which we apply three-step $\ell_\infty$ PGD ($\epsilon=8/255$ and step size $2/255$) to all camera streams. In contrast to the original backbone-specific training recipes, we use the native LeRobot~\citep{cadene2024lerobot} loss of each evaluated VLA and fine-tune all five backbones using LoRA with rank 32. State observations and output actions are not perturbed during this training, and we evaluate the checkpoint after 8,000 updates.

\paragraph{VLA-Corrector.}
This method predicts the short-horizon change in visual features and interrupts an action chunk when the predicted and observed changes become persistently inconsistent. Using the authors' released implementation, we train a separate residual MLP corrector for each backbone on frozen visual features extracted from clean LIBERO demonstrations. We use a history window of one step, a prediction interval and nominal execution length of 10 steps, and train each corrector for 30 epochs with the cosine loss. At inference, we use guidance strength $\eta=1$ and set the post-intervention cooldown to zero. We evaluate $\pi_0$, $\pi_{0.5}$, SmolVLA, and X-VLA. We exclude $\pi_0$-FAST because its autoregressive action generation is incompatible with the method's flow-guidance step.

\paragraph{AAC.}
This method selects an execution length from the entropy profile of action chunks sampled repeatedly for the same observation. We draw 20 chunks, compute Gaussian differential entropy for translation and rotation and binary entropy for the gripper, and apply the original maximum-entropy-difference rule with the minimum-action-magnitude threshold $\alpha=3$. Entropy and action magnitude are computed after the policy and environment action postprocessing so that they are measured in the executable LIBERO action space. Because LIBERO represents rotation as a three-dimensional delta rotation vector, we accumulate these vectors instead of composing quaternions when computing the rotation magnitude. To apply AAC to the otherwise deterministic autoregressive $\pi_0$-FAST policy, we generate 10-step candidate chunks using stochastic token sampling with temperature $0.7$; the other models use their native stochastic action samplers.

\subsection{Proposed Method Implementation}\label{app:proposed_method_implementation}
In this section, we describe the implementation details of our proposed method.
We provide a pseudocode in Algorithm~\ref{alg:risk_based_action_chunking}.
CARE accumulates the discounted action discrepancy from the start of the aligned sequences.
It selects the longest prefix whose cumulative discrepancy does not exceed $\tau$, subject to the execution-length cap.
If the first action already exceeds the threshold, CARE executes one action before the next inference step.
We use the discrepancy function defined in Section~\ref{sec:method}, with a small constant $\epsilon$ in the denominator to avoid division by zero.
At the start of each episode, no previous action chunk is available.
We therefore set the initial execution length to $H_{\mathrm{init}}=1$.
Appendix~\ref{app:cold_start_analysis} evaluates the effect of this initialization.
For the LIBERO experiments, we set the discount factor to $\gamma=0.85$ and cap the execution length at 15 steps.
We use $\tau=1.5$ for $\pi_0$, $\pi_0$-FAST, and SmolVLA, and $\tau=1.0$ for $\pi_{0.5}$ and X-VLA.

\begin{algorithm}[t]
    \caption{CARE: Adaptive Execution Length Selection}
    \label{alg:risk_based_action_chunking}
    \begin{algorithmic}[1]
        \Require Current length-$K$ action chunk $\mathbf A_t$,
        preceding length-$K$ chunk $\mathbf A_{\mathrm{prev}}$,
        preceding execution length $H_{\mathrm{prev}}\in\{1,\ldots,H_{\max}\}$.
        \Ensure Execution length $H_t\in\{1,\ldots,H_{\max}\}$

        \If{$\mathbf A_{\mathrm{prev}}=\varnothing$}
            \State $H_t\gets H_{\mathrm{init}}$
        \Else
            \State $\widehat{\mathbf A}\gets
                \mathbf A_{\mathrm{prev}}^{H_{\mathrm{prev}}+1:H_{\mathrm{prev}}+H_{\max}}$
            \State $R\gets0,\ H_t\gets1$
            \For{$i=0$ to $H_{\max}-1$}
                \State
                $R\gets R+\gamma^i
                \dfrac{
                    \lVert\mathbf A_t^{i+1}-\widehat{\mathbf A}^{i+1}\rVert_2
                }{
                    \lVert\mathbf A_t^{i+1}\rVert_2+
                    \lVert\widehat{\mathbf A}^{i+1}\rVert_2+\epsilon
                }$
                \If{$R>\tau$}
                    \State \textbf{break}
                \EndIf
                \State $H_t\gets i+1$
            \EndFor
        \EndIf
        \State \Return $H_t$
    \end{algorithmic}
\end{algorithm}

\subsection{Real-World Experiment Details}\label{app:real_world_experiment_details}
In this section, we provide additional details about the real-world experiments described in Section~\ref{subsec:real_world_robot_experiments}.

\begin{figure}[t]
    \centering
    \includegraphics[width=\linewidth]{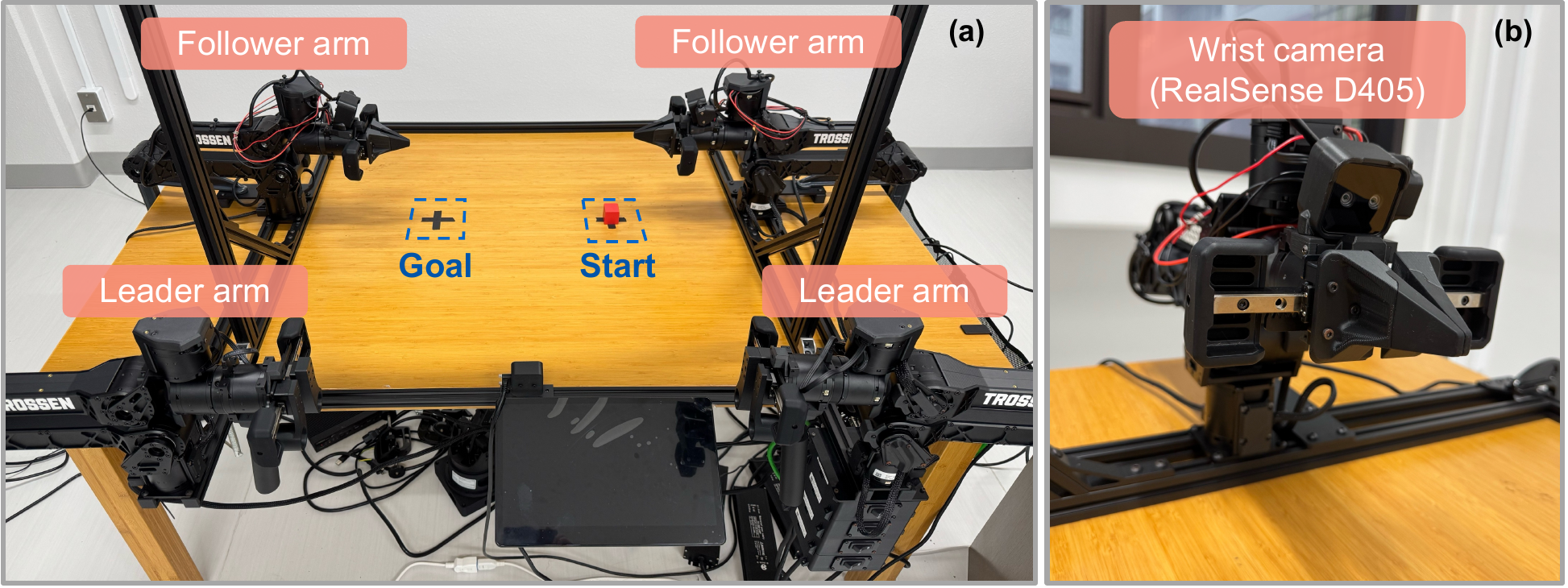}
    \caption{\textbf{Real-world experimental setup.}
    (a) Workspace overview showing the two follower arms and the leader arms used for teleoperated demonstration collection. The blue dashed boxes mark the block's start and goal locations and are annotations added to the image.
    (b) Follower arm with a wrist-mounted Intel RealSense D405 camera.}
    \label{fig:real_world_setup}
\end{figure}

\paragraph{Hardware Setup.}
We deploy the VLA on two 6-DoF WidowX AI follower arms equipped with grippers, as shown in Figure~\ref{fig:real_world_setup}.
Visual inputs are provided by four Intel RealSense D405 cameras. Two external cameras are mounted at upper and lower positions on the frame, and one wrist-mounted camera is on each arm.
% We design a pick-and-place task to assess real-world performance: 右アームでブロックを把持して左アームに受け渡し、左アームで所定の位置に配置する。
% 各エピソードの実行時間は30秒とし、学習済みモデルは通常約20秒でタスクを完了した。

\paragraph{Task Design.}
We design a pick-and-place task to assess real-world performance.
In this task, the right arm grasps a red block and hands it to the left arm, which places it at a designated location.
An episode is considered successful if visual inspection confirms that the red block has been placed at the designated location after the handover.
Each episode lasts 40 seconds, and the trained model typically completes the task in approximately 20 seconds, including inference waiting time.

% ベースモデルには$\pi_{0.5}$を用い、100エピソードのデモンストレーションからLoRAによるファインチューニングを40,000ステップ行った。
% LoRAのrankとalphaはいずれも32、バッチサイズは16とし、最適化には学習率$2.5\times10^{-5}$のAdamWを用いた。
% 行動チャンクの実行長は$H=25$とした。
% 学習済みモデルは、摂動のない条件で実施した15エピソードにおいて100\%の成功率を達成した。

\paragraph{Model Training.}
We fine-tune $\pi_{0.5}$ using LoRA for 40,000 steps on 100 demonstration episodes.
Both the LoRA rank and alpha are set to 32, and we use a batch size of 16 and AdamW with a learning rate of $2.5\times10^{-5}$.
The action chunk execution length is set to $H=25$.
Input perturbations caused unsafe movements in preliminary experiments. 
To mitigate this, we clip the predicted joint-position targets to the joint-specific SDK position limits and constrain each revolute-joint target to within ±0.10 rad of its current, unperturbed measured position before execution.
These constraints are applied at every control step in all evaluations, both with and without input perturbations.

% 予備実験では、入力摂動によって危険な動作が生じたため、安全のためにモデルの出力する［制約対象］を［制約範囲］に制限した。
% この出力制約は、摂動の有無によらずすべての評価に適用した。

\paragraph{Perturbation Design.}

% 摂動には、data loss、random state、blurの3種類を用いた。条件を統制して評価するため、ロボットや周囲の環境に物理的な外乱を与える代わりに、推論時の入力に人工的な摂動を適用した。

We evaluate three perturbation types: data loss, random state, and blur.
To control the evaluation conditions, we apply synthetic perturbations to the inputs at inference time rather than physically disturbing the robot or its environment.
% すべての推論時刻を個別に評価するには多くの実機試行を要するため、シミュレーション実験と同様に、エピソードを複数の区間に分けて摂動の適用時刻を選択した。
% 具体的には、摂動のない実行におけるロボットの動作に基づき、タスクを次の4つのphaseに分割した（Figure~3）。
% (i) 物体の探索：右アームがカメラでブロックを捉える段階、
% (ii) 物体の把持：右アームがブロックを把持して持ち上げる段階、
% (iii) 持ち替え：右アームから左アームにブロックを受け渡す段階、
% (iv) 物体の配置：左アームがブロックを所定の位置に配置する段階である。
% 各phaseは100制御ステップからなる固定の時間区間とし、すべての評価エピソードで共通の区間を用いた。
Evaluating every inference step individually would require many real-world trials, so we follow the simulation protocol and divide each episode into intervals within which perturbation timing is selected.
Based on the robot's behavior during unperturbed execution, we divide the task into four phases (Figure~\ref{fig:real_world_failure_examples}):
(i) search, in which the right arm brings the block into camera view;
(ii) grasping, in which the right arm grasps and lifts the block;
(iii) handover, in which the right arm transfers the block to the left arm; and
(iv) placement, in which the left arm places the block at the designated location.
Each phase is a fixed interval of 100 control steps, and the same intervals are used across all evaluation episodes.
% 各評価エピソードでは、対象とするphaseを一つ定め、その区間内の推論時刻をランダムに一つ選択した。
% 選択した1回の推論に渡す入力にのみ摂動を適用し、それ以外の推論では正常な入力を用いた。
% したがって、各エピソードにおける摂動の適用は1回のみであり、異なるphaseは別々のエピソードで評価した。
% 各摂動・各phaseについて15エピソードから成功率を算出した。
We compute the success rate over 15 episodes for each perturbation type and phase.

\section{Additional Experiments}\label{app:additional_experiments}

\subsection{Perturbation Timing Analysis}\label{app:perturbation_timing_analysis}
\begin{figure}[t]
    \centering
    \includegraphics[width=\linewidth]{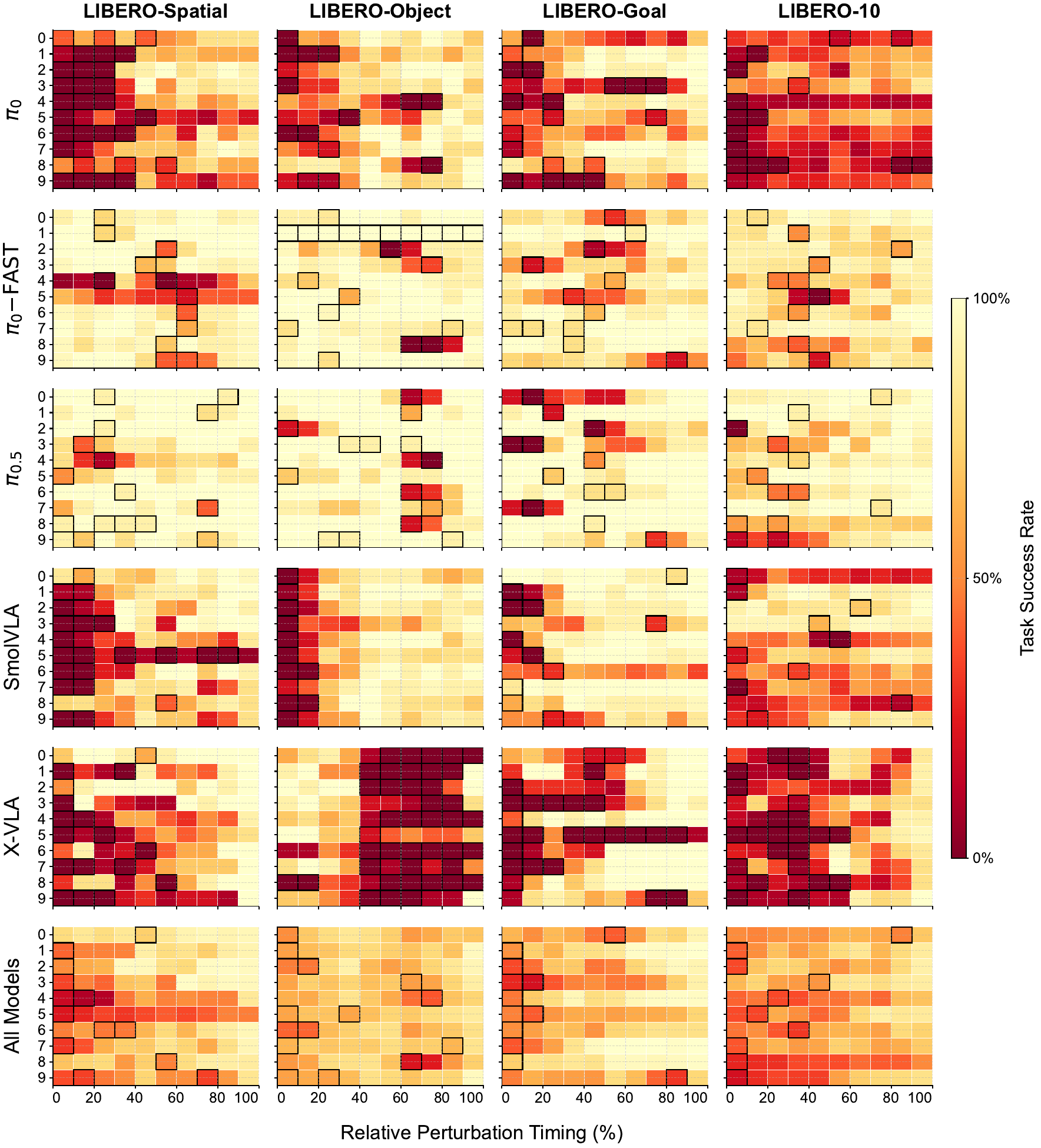}
    \caption{\textbf{Perturbation timing analysis by model and suite.} Rows show $\pi_0$, $\pi_0$-FAST, $\pi_{0.5}$, SmolVLA, X-VLA, and their equal-weight average. Columns show LIBERO-Spatial, LIBERO-Object, LIBERO-Goal, and LIBERO-10.}
    \label{fig:perturbation_timing_analysis_all_suites}
\end{figure}

% 本セクションでは、Section4.2におけるperturbation timingの影響について、より広範な評価を行います。
% 私たちは、各モデル、suiteにおけるtiming毎の成功率をFigure13に示します。
This section extends the evaluation of perturbation timing in Section~\ref{subsec:eval_vulnerability_finding1} to all four LIBERO suites.
Figure~\ref{fig:perturbation_timing_analysis_all_suites} shows task success rates under one-step data loss across perturbation timings for each model and suite.
We group relative perturbation timings into ten bins.

% 摂動のタイミングの影響
\paragraph{Effect of perturbation timing.}
% 多くの設定において、すべてのtimingではなく、特定のタイミングにおいて特に大きい性能低下が起きていることが確認できます。例えば、pi05では、わずか10%のrelative timingでは0に近い成功率を示す一方で、それ以外のtimingではほとんど成功率が低下していないtaskが存在しています。また、モデルによってどれくらいの範囲が脆弱であるかは変化しています。pi05, pi0-fastでは、非常に短い区間でのみ性能低下が発生する一方で、他のモデルではより長い区間で性能低下が起こっている傾向が確認できます。
In many settings, large performance drops occur at specific perturbation timings rather than uniformly across all timings.
For example, on some tasks, $\pi_{0.5}$ has near-zero success rates within a single 10\% relative timing interval.
At other timings, the success rate on these tasks shows little decline.
The range of vulnerable timings also varies across models.
Performance drops for $\pi_{0.5}$ and $\pi_0$-FAST tend to be confined to very short intervals.
Other models tend to show performance drops over wider intervals.

% 原因分析
\paragraph{Analysis of possible causes.}
% 私たちは、この現象に対して2つの理由があると考えています。1つ目は、taskにはcriticalなタイミングが存在すると言うことです。例えば、LIBEROではアームを用いて物体を運ぶタスクが存在しますが、物体を把持する時や、置く時など、特にsensitiveなタイミングが存在し、そのようなタイミングで摂動が与えられると、大きな影響を及ぼすと考えられます。2つ目は、compounding errorです。全体の傾向として、序盤の雪道の方が大きな影響を与えていることが確認できます。これは、compounding errors, a well-known problem in imitation learning~\citep{ross2011reduction,xu2024humanvla}: an early erroneous action shifts subsequent states away from the demonstration distribution, increasing the likelihood of further errors throughout the remaining rollout.
% こうした結果から、特に脆弱なタイミングを事前に見つけたり、そのようなタイミングで攻撃する手法の開発などにつながると考えています。
We consider two possible explanations for these patterns.
First, tasks contain critical moments that are particularly sensitive to perturbations.
For example, some LIBERO tasks require a robot arm to transport objects.
In such cases, perturbations may have a particularly large impact when the robot grasps or places an object.
Second, compounding errors may explain why early perturbations tend to have a greater impact.
Compounding errors are a well-known problem in imitation learning~\citep{ross2011reduction,xu2024humanvla}.
An early erroneous action can shift subsequent states away from the demonstration distribution.
This shift increases the likelihood of further errors throughout the remaining rollout.
These findings motivate methods to identify vulnerable timings in advance and attacks that target such timings.

\subsection{Experiments on Meta-World Benchmark}\label{app:meta_world_experiments}
We conduct experiments on Meta-World to evaluate vulnerability to one-step perturbations and the generalization of CARE beyond LIBERO. Meta-World is a simulated benchmark with 50 robotic manipulation tasks~\citep{yu2020metaworld}. These tasks involve a Sawyer robot arm and everyday objects in a shared tabletop environment. We evaluate on all 50 tasks in the MT50 suite.

\paragraph{Setup.}
We fine-tune $\pi_0$, $\pi_{0.5}$, and SmolVLA on the Meta-World MT50 demonstration dataset. We use the checkpoints at 30,000 training steps for all three models and a global batch size of 32. The peak learning rate is $2.5\times10^{-5}$ for $\pi_0$ and $\pi_{0.5}$, and $1\times10^{-4}$ for SmolVLA. All models use 1,000 warmup steps. Vanilla uses a fixed execution length of $H=15$. CARE uses $\tau=1.5$ and $\gamma=0.85$, with a maximum execution length of 15 and a cold-start execution length of 1. Each episode has a maximum of 300 environment steps.
We evaluate 10 episodes per task for the clean condition and each perturbation position.

\begin{figure}[t]
    \centering
    \includegraphics[width=\linewidth]{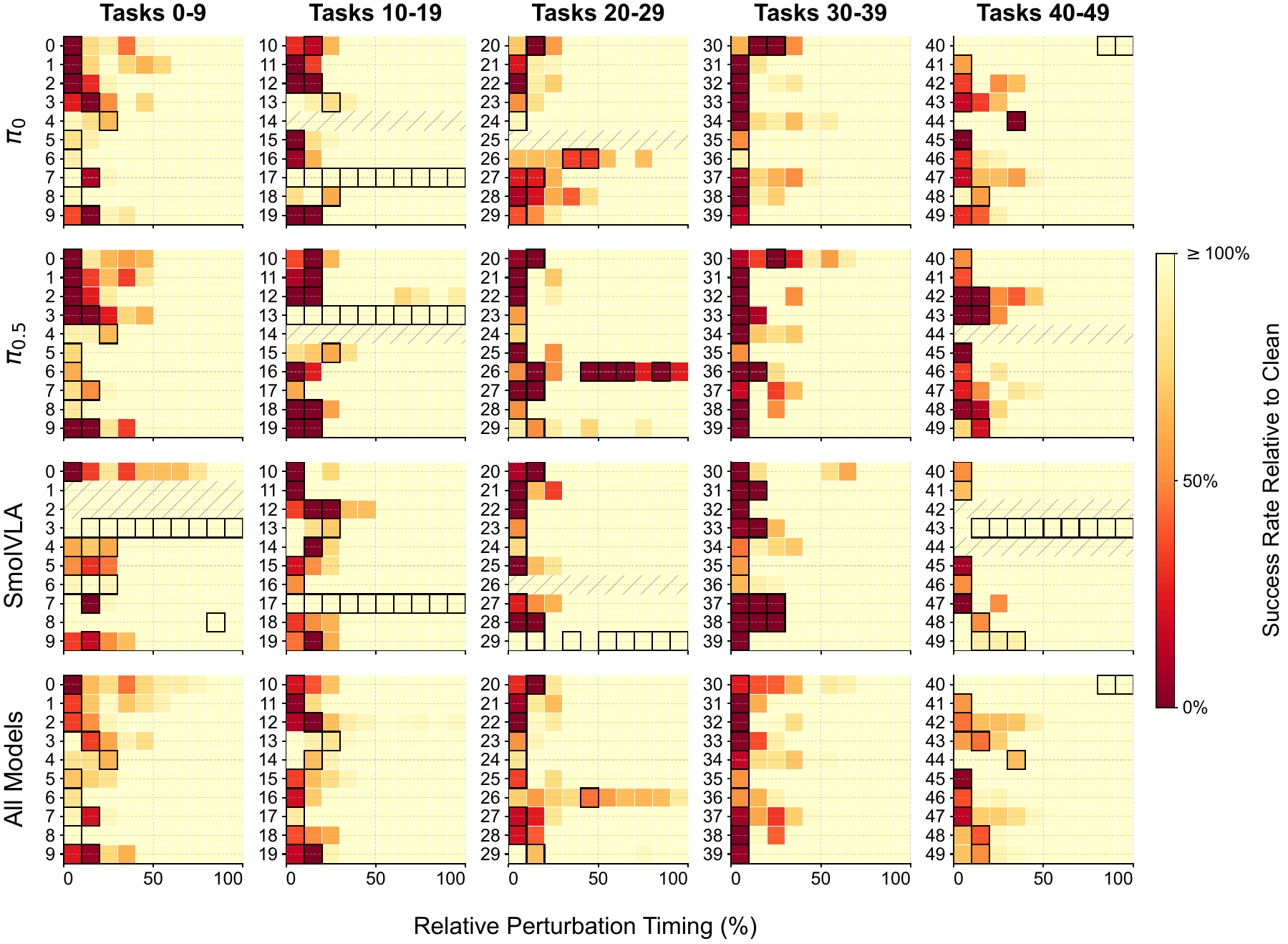}
    \caption{\textbf{Perturbation timing analysis on Meta-World MT50.} Values show success rates under data loss relative to clean performance. Hatching indicates tasks with zero clean success rate. Black boxes mark the timing bins with the lowest relative success rate for each task.}
    \label{fig:metaworld_perturbation_timing}
\end{figure}

\paragraph{Effect of perturbation timing.}
We first analyze how perturbation timing affects success rates on Meta-World. Figure~\ref{fig:metaworld_perturbation_timing} shows the results.
Some tasks in Meta-World have low success rates even under clean conditions.
We therefore use success rates relative to the clean success rate of each task. Hatching marks tasks with zero clean success rates. We exclude these tasks from this analysis.
The effect of a perturbation varies greatly across steps. This pattern is similar to that in LIBERO. However, vulnerability is more concentrated at early steps in Meta-World than in LIBERO.
These results suggest that perturbation timing may be important across different tasks and environments. However, the environment may strongly influence which steps are vulnerable to perturbations.

\begin{table}[t]
    \centering
    \caption{Clean success rate and Bottom10\%-SR (\%, $\uparrow$) under one-step observation perturbations on Meta-World MT50. Each task receives 10 episodes per condition. Bold and underlining indicate the highest and second-highest scores within each model, respectively.}
    \label{tab:metaworld_robustness}
    \begingroup
    \small
    \setlength{\tabcolsep}{5pt}
    \renewcommand{\arraystretch}{1.08}
    \resizebox{0.95\textwidth}{!}{%
    \begin{tabular}{llc>{\columncolor{imageinputbg}}c>{\columncolor{stateinputbg}}c>{\columncolor{bothinputbg}}c}
        \toprule
        \multirow{2}{*}{Model} & \multirow{2}{*}{Method} & \multirow{2}{*}{Clean} & \multicolumn{3}{c}{Bottom10\%-SR} \\
        \cmidrule(lr){4-6}
        & & & Blur & Random state & Data loss \\
        \midrule
        \multirow{5}{*}{$\pi_0$}
          & Vanilla       & 71.6 & 56.9 & 29.4 & 23.5 \\
        \cdashline{2-6}
          & RobustVLA     & 47.2 & 41.0 & 22.6 & 21.6 \\
          & VLA-Corrector & \underline{74.2} & 59.4 & 32.6 & 28.4 \\
          & AAC           & 74.0 & \underline{62.7} & \underline{45.5} & \underline{42.4} \\
          & \cellcolor{black!7}\textbf{CARE (ours)} & \cellcolor{black!7}\textbf{76.4} & \cellcolor{imageinputbg!93!black}\textbf{63.0} & \cellcolor{stateinputbg!93!black}\textbf{59.6} & \cellcolor{bothinputbg!93!black}\textbf{58.2} \\
        \midrule
        \multirow{5}{*}{$\pi_{0.5}$}
          & Vanilla       & 62.0 & 50.5 & 32.0 & 17.5 \\
        \cdashline{2-6}
          & RobustVLA     & 40.6 & 37.7 & 18.1 & 11.3 \\
          & VLA-Corrector & 61.2 & 50.4 & 34.2 & 19.6 \\
          & AAC           & \underline{63.8} & \underline{50.9} & \underline{38.5} & \underline{31.0} \\
          & \cellcolor{black!7}\textbf{CARE (ours)} & \cellcolor{black!7}\textbf{68.4} & \cellcolor{imageinputbg!93!black}\textbf{56.1} & \cellcolor{stateinputbg!93!black}\textbf{48.3} & \cellcolor{bothinputbg!93!black}\textbf{39.7} \\
        \midrule
        \multirow{5}{*}{SmolVLA}
          & Vanilla       & \underline{49.4} & \textbf{47.0} & 21.5 & 21.7 \\
        \cdashline{2-6}
          & RobustVLA     & 41.0 & 39.3 & 18.0 & 16.8 \\
          & VLA-Corrector & \underline{49.4} & 45.4 & 22.8 & 23.8 \\
          & AAC           & 49.2 & 44.5 & \underline{31.8} & \underline{28.1} \\
          & \cellcolor{black!7}\textbf{CARE (ours)} & \cellcolor{black!7}\textbf{51.2} & \cellcolor{imageinputbg!93!black}\underline{46.8} & \cellcolor{stateinputbg!93!black}\textbf{37.7} & \cellcolor{bothinputbg!93!black}\textbf{36.2} \\
        \bottomrule
    \end{tabular}%
    }
    \endgroup
\end{table}

\paragraph{One-step vulnerability persists on Meta-World.}
Table~\ref{tab:metaworld_robustness} reports the results. One-step perturbations reduce the performance of Vanilla for all three models. Data loss causes particularly large drops, from clean success rates of 71.6\%, 62.0\%, and 49.4\% to 23.5\%, 17.5\%, and 21.7\%, respectively. Random-state perturbations also cause larger drops than motion blur for all three models. This pattern differs from the LIBERO results for SmolVLA, where image perturbations have a greater effect than state perturbations. These results suggest that vulnerability across input modalities depends on the environment as well as the model.

\paragraph{CARE remains effective on Meta-World.}
CARE achieves the highest clean success rate and the highest Bottom10\%-SR under random-state and data-loss perturbations for all three models. It also achieves the highest motion-blur score for $\pi_0$ and $\pi_{0.5}$. For SmolVLA, CARE is slightly below Vanilla under motion blur, at 46.8\% versus 47.0\%. These results support the effectiveness of CARE beyond LIBERO. However, CARE still exhibits substantial gaps between clean and perturbed performance on Meta-World, in contrast to the near-clean performance observed in several LIBERO settings. This suggests that adaptive execution length selection alone may not fully address one-step perturbations in this benchmark. A combination of CARE and complementary recovery methods may help close the remaining gap.

\begin{figure}[t]
    \centering
    \includegraphics[width=\linewidth]{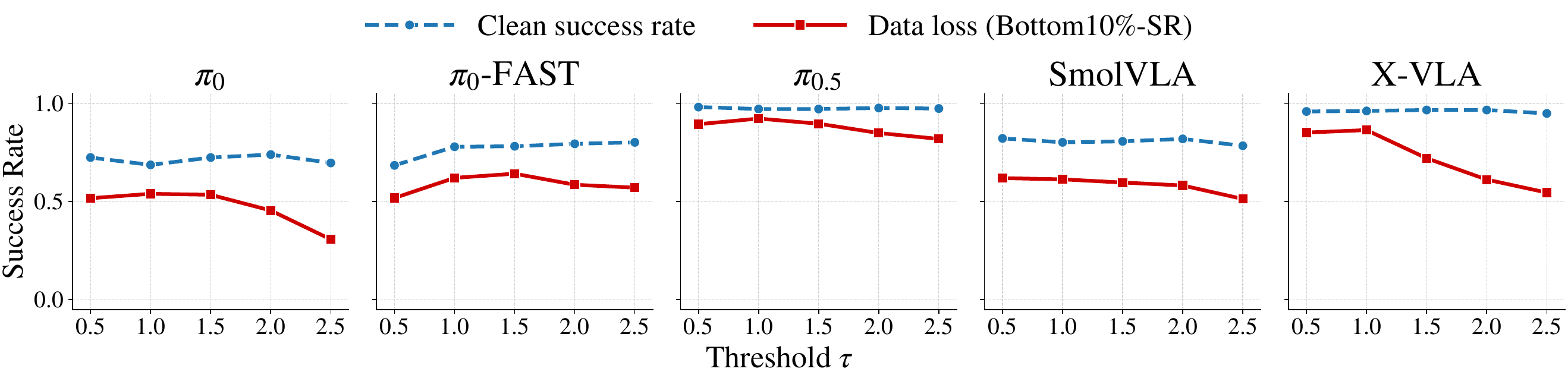}
    \caption{\textbf{Sensitivity to the discrepancy threshold $\tau$.} Each panel shows clean success rate and Bottom10\%-SR under data loss, averaged over 40 tasks across the four LIBERO suites.}
    \label{fig:hyperparameter_sensitivity}
\end{figure}

\subsection{Hyperparameter Sensitivity}
We analyze the sensitivity of CARE to the discrepancy threshold $\tau$. This threshold controls the cumulative discounted discrepancy that CARE permits between consecutive action predictions. A smaller $\tau$ makes CARE more sensitive to changes in these predictions and favors shorter execution lengths. Shorter execution lengths can improve robustness to one-step perturbations, but they require more frequent policy inference. The choice of $\tau$ therefore determines the trade-off between robustness and inference cost.

Figure~\ref{fig:hyperparameter_sensitivity} shows the results for $\tau$ from 0.5 to 2.5 in increments of 0.5. We fix $\gamma=0.85$, the maximum execution length at 15, and $H_{\mathrm{init}}=1$. Clean performance remains relatively stable for most models, whereas robustness to data loss generally decreases at larger thresholds. A larger $\tau$ allows longer execution lengths even under perturbations, so erroneous actions can affect the environment for more steps. This behavior is consistent with the lower robustness at larger thresholds. Within the range of 0.5--1.5, robustness varies relatively little for $\pi_0$, $\pi_{0.5}$, and SmolVLA. However, X-VLA shows a larger decrease at $\tau=1.5$, and $\pi_0$-FAST has lower clean performance at $\tau=0.5$. These results suggest that CARE does not require precise threshold tuning for several models, although the appropriate range depends on the model.

\begin{figure}[t]
    \centering
    \includegraphics[width=\linewidth]{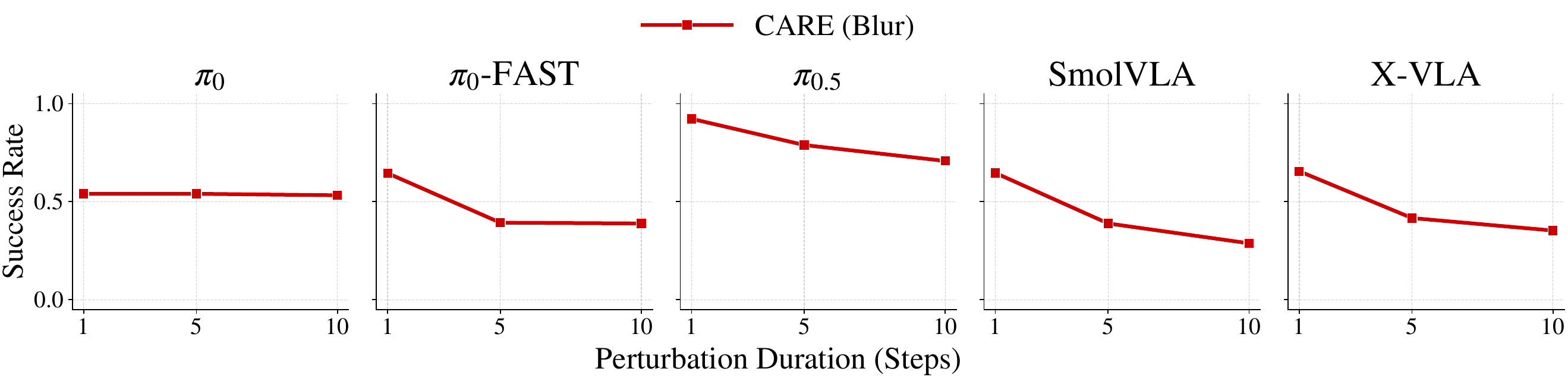}
    \caption{\textbf{Robustness to multi-step motion blur.} Red curves show Bottom10\%-SR under Blur at durations of 1, 5, and 10 steps. Success rate is averaged over 40 tasks across the four LIBERO suites, with 10 trials.}
    \label{fig:multistep_blur}
\end{figure}

\begin{figure}[t]
    \centering
    \includegraphics[width=\linewidth]{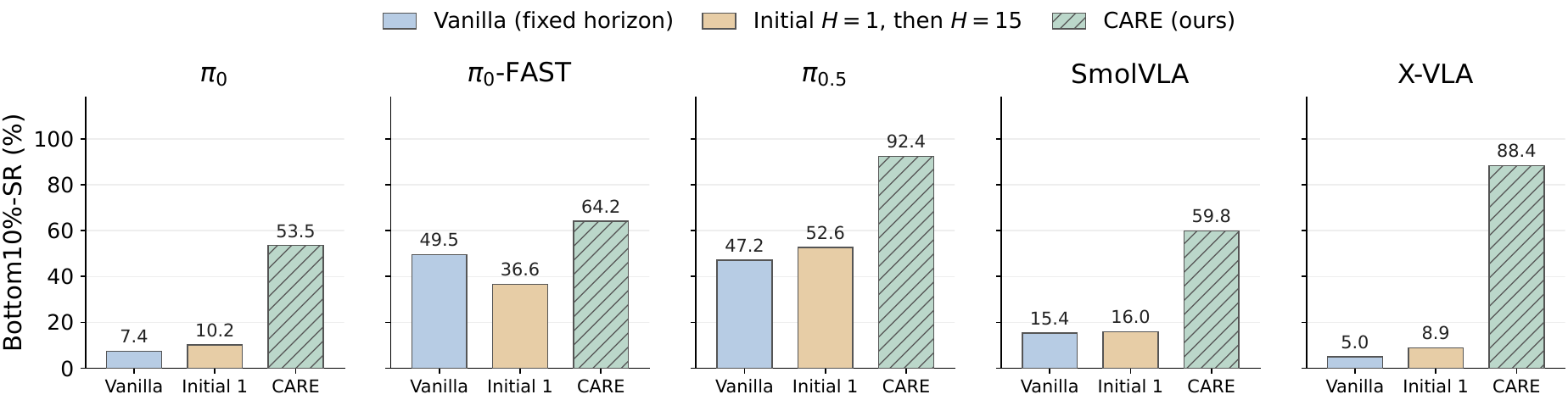}
    \caption{\textbf{Effect of cold start in CARE.} Comparison of Vanilla (fixed execution length), an initial-$H=1$ variant with $H=15$ thereafter, and CARE under data loss. The cold-start-only evaluation covers 40 tasks across four LIBERO suites, with 10 episodes per setting.}
    \label{fig:cold_start_ablation}
\end{figure}

\subsection{Robustness to multi-step perturbations.}
\label{app:multistep_perturbations}

% 本セクションでは、multi-step perturbationに対する頑健性を評価する。本評価では、controlled experimentのためにone-step perturbationに焦点を当てました。しかしながら、現実の設定ではより複数のステップにおいて観測がcorruptedされることがありえます。そこで、私たちはperturbationを与えるstep数を1より長く設定します。具体的には、5stepおよび10step連続で摂動が与えられた時のCAREの有効性を評価します。
We evaluate robustness to multi-step perturbations in this section.
Our main evaluation focuses on one-step perturbations to ensure controlled experimental conditions.
However, observations in real-world settings may be corrupted over multiple consecutive steps.
We therefore extend the perturbation duration beyond a single step and evaluate CARE under perturbations that last for 5 and 10 consecutive steps.

% Figure11に結果をしまします。stepが増加するにつれて、CAREの有効性が低下していることが確認できます。これは、長時間において観測が摂動されると、短い実行長を選択しても、次の推論でも観測がcorrputedされているため、正常な行動をreplanできないためです。このことから、実効長選択を入れ換える方法ではmulti-step perturbationに対する完全な頑健性を得ることには限界があり、recoveryなどと組み合わせることが重要になります。
Figure~\ref{fig:multistep_blur} shows that CARE becomes less effective as the perturbation duration increases.
When corruption persists, observations remain corrupted at the next inference step even if CARE selects a short execution length.
The policy therefore cannot replan its actions from clean observations.
These results show that execution length selection alone has limited ability to provide full robustness to multi-step perturbations.
It is therefore important to combine CARE with complementary methods, such as recovery mechanisms.

\subsection{Analysis of Cold Start in CARE}
\label{app:cold_start_analysis}
% CAREでは、最初の推論における比較対象となるaction chunkを用意するために、Cold Startを適用しています。我々の調査では、特に初期のステップにおける摂動が大きな性能低下を引き起こすことがわかっています。そのため、本セクションはこのcold startが顔牽制にどのような影響を及ぼすのかを調査します。私たちは、ベースラインとして、最初の実効長だけをH=1として、それ以降をH=15とする推論方式, initial 1を用います。この推論方式に対してvanillaとCAREの頑健性を評価することで、cold startの影響を評価します。
CARE uses a cold start because no previous action chunk is available for comparison at the first inference step.
It executes one action from the first chunk and uses the remaining actions as a reference at the next inference step.
Our analysis shows that perturbations at early steps can cause particularly large performance drops.
We therefore investigate how this cold start affects robustness.
We introduce a baseline, \textit{initial 1}, that sets $H=1$ at the first inference step and $H=15$ at every subsequent inference step.
We compare its robustness with that of Vanilla and CARE to assess the effect of the cold start.

% 実験結果をFigure12に示します。initial 1はCAREと比較して十分な顔牽制の向上を示しませんでした。多くのモデルにおいてVanillaと同等の成功率を示しました。この結果は、cold startだけでは、one-step perturbationに対する頑健性を十分に獲得できていなことを示しています。したがって、CAREの頑健性向上は、cold startによるものではなく、動的な実効長選択によって摂動下によって短い実行長を選択できているためであることがわかります。
Figure~\ref{fig:cold_start_ablation} reports the results under one-step data loss.
The \textit{initial 1} baseline provides much smaller robustness gains than CARE and achieves success rates comparable to Vanilla for many models.
These results show that the cold start alone is insufficient to achieve robustness to one-step perturbations.
CARE's robustness gains instead come from adaptive execution length selection, which allows it to select short execution lengths under perturbations.

\end{document}